\documentclass[acmsmall]{acmart}

\AtBeginDocument{%
  }

\setcopyright{acmcopyright}
\copyrightyear{2023}
\acmYear{2023}
\acmDOI{XXXXXXX.XXXXXXX}

\acmJournal{JACM}
\acmVolume{37}
\acmNumber{4}
\acmArticle{111}
\acmMonth{8}

\usepackage{booktabs}                                  % for nice tables
\usepackage{bbding,pifont}                             % for the check mark and cross sign
\usepackage{multirow}                         		   % for table cell taking multiple rows
\usepackage{algorithm,algpseudocode}  				   % For algorithms
\usepackage{bm}
\usepackage{xcolor}
\usepackage{amsmath}

\usepackage{cleveref}
\usepackage{caption}
\usepackage{color}
\usepackage{graphicx}
\usepackage{pifont}

\usepackage{subcaption}

\newcommand{\ourmethod}{{DFC-Net}}
\newcommand{\mixamodata}{\textsc{Mixamo-Pose}}
\newcommand{\edndata}{\textsc{EDN-10k}}
\newcommand{\supdata}{\textsc{Mixamo-Sup}}

\newcommand{\sfe}{Static Feature Encoder}
\newcommand{\mfe}{Pose Feature Encoder}
\newcommand{\gen}{Image Generator}
\newcommand{\dis}{Image Discriminator}
\newcommand{\pe}{Pose Estimator}
\newcommand{\ka}{Keypoint Amplifier}
\newcommand{\mfr}{Pose Refiner}

\newcommand{\field}[1]{\mathbb{#1}}
\newcommand{\R}{\field{R}}                              % real domain

\newcommand{\mat}[1]{\boldsymbol{#1}}

\newcommand{\bx}        {\mat{x}}

\newcommand{\bh}        {\mat{h}}
\newcommand{\bpaf}      {\mat{p}}

\newcommand{\rt}{\mathrm{t}}
\newcommand{\rs}{\mathrm{s}}
\newcommand{\rsyn}{\mathrm{syn}}
\newcommand{\radv}{\mathrm{adv}}
\newcommand{\rper}{\mathrm{per}}
\newcommand{\rfm}{\mathrm{fm}}
\newcommand{\rmc}{\mathrm{mc}}
\newcommand{\rsc}{\mathrm{sc}}
\newcommand{\rfull}{\mathrm{full}}
\newcommand{\rsup}{\mathrm{sup}}
\newcommand{\rtrain}{\mathrm{tr}}

\newcommand{\mL}        {\mathcal{L}}

\newcommand{\dif}[1]{#1}
\newcommand{\diff}[1]{#1}

\newcommand{\eat}[1]{}                                 % to hide contents easily
\begin{document}

\title{Human Pose Transfer with Augmented Disentangled Feature Consistency}

\author{Kun Wu}
\email{kwu102@syr.edu}
\affiliation{%
  \institution{Syracuse University}
  \city{Syracuse}
  \state{NY}
  \country{USA}
  \postcode{13244}
}

\author{Chengxiang~Yin}
\email{cyin02@syr.edu}
\affiliation{%
  \institution{Syracuse University}
  \city{Syracuse}
  \state{NY}
  \country{USA}
  \postcode{13244}
}

\author{Zhengping Che}
\email{chezp@midea.com, chezhengping@gmail.com}
\affiliation{%
  \institution{Midea Group}
  \city{Beijing}
  \country{China}
}

\author{Bo Jiang}
\email{scottjiangbo@didiglobal.com}
\affiliation{%
 \institution{Didi Chuxing}
 \city{Beijing}
 \country{China}}

\author{Jian Tang$^{\dagger}$}
\authornote{$^\dagger$Corresponding author: Jian Tang.}
\email{tangjian22@midea.com}
\affiliation{%
  \institution{Midea Group}
  \city{Beijing}
  \country{China}
}

\author{Zheng Guan}
\email{guanzheng\_bit@hotmail.com}
\affiliation{%
  \institution{Computer Science School, Beijing Institute of Technology}
  \city{Beijing}
  \country{China}
}

\author{Gangyi Ding}
\email{dinggangyi\_bit@hotmail.com}
\affiliation{%
  \institution{Computer Science School, Beijing Institute of Technology}
  \city{Beijing}
  \country{China}
}

\renewcommand{\shortauthors}{Wu et al.}

\begin{abstract}
    Deep generative models have made great progress in synthesizing images with arbitrary human poses and transferring the poses of one person to others.
Though many different methods have been proposed to generate images with high visual fidelity, the main challenge remains and comes from two fundamental issues: pose ambiguity and appearance inconsistency.
To alleviate the current limitations and improve the quality of the synthesized images, 
we propose a pose transfer network with augmented \textbf{D}isentangled \textbf{F}eature \textbf{C}onsistency ({\ourmethod}) to facilitate human pose transfer.
Given a pair of images containing the source and target person, {\ourmethod} extracts pose and static information from the source and target respectively, then synthesizes an image of the target person with the desired pose from the source.
Moreover, {\ourmethod} leverages disentangled feature consistency losses in the adversarial training to strengthen the transfer coherence and integrates a keypoint amplifier to enhance the pose feature extraction.
With the help of the disentangled feature consistency losses, we further propose a novel data augmentation scheme that introduces unpaired support data with the augmented consistency constraints to improve the generality and robustness of {\ourmethod}.
Extensive experimental results on {\mixamodata} and {\edndata} have demonstrated {\ourmethod} achieves state-of-the-art performance on pose transfer.
\end{abstract}



\received{11 August 2022}
\received[revised]{9 August 2023}
\received[accepted]{9 September 2023}

\maketitle

\section{Introduction}
\label{sec:intro}
% Human pose transfer has become increasingly compelling recently since it is closely related to movies' special effects~\cite{tung2017self}, reenactment~\cite{Liu2019neural}, person re-identification~\cite{Liu2020Pair,Tian2021}, entertainment systems~\cite{Xia2017} and so forth~\cite{Moeslund2006,Ding2016}.
\dif{
Human pose transfer has become increasingly compelling recently since it can be applied to real-world applications such as movies' special effects~\cite{tung2017self}, entertainment systems~\cite{Xia2017}, reenactment~\cite{Liu2019neural} and so forth~\cite{Moeslund2006,Ding2016}.
At the same time, it is also closely related to many computer vision tasks like human-object interaction recognition~\cite{Qi2018learning,Zhou2020cascaded,Zhou2021cascaded}, 
person re-identification~\cite{Liu2020Pair,Tian2021}, human pose segmentation~\cite{Zhou2022consistency,Wang2020paying} and human parsing~\cite{Li2020self, Wang2021hierarchical}, and all these methods can be beneficial to each other.
}
Given some images of a target person and a source person image with the desired pose (e.g., judo, dance), the goal of the human pose transfer task is to synthesize a realistic image of the target person with the desired pose of the source person.

With the power of deep learning, especially the generative adversarial networks (GANs)~\cite{Goodfellow2014}, pioneering works
% have raised impressive solutions to address the human image generation~\cite{Mo2019,Liu2019SwapGAN,Gu2021,Zhan2021} by efficiently leveraging the image-to-image translation schemes and have achieved significant progress. 
have raised impressive solutions to address the human image generation~\cite{Ma2017,Neverova2018,Liu2019} by efficiently leveraging the image-to-image translation schemes and have achieved significant progress. 
Intuitively, early routine coarsely conducts human pose transfer through general image-to-image translation methods such as Pix2Pix~\cite{Isola2017} and CycleGAN~\cite{Zhu2017}, which attempt to translate the extracted skeleton image of \dif{the} source person to the image of target person with the desired poses.

Subsequent approaches~\cite{Ma2017,Ma2018,Li2019} adopt specifically designed modules for human pose transfer.
Specifically, the U-net architecture with skip connections in~\cite{Esser2018} is employed to keep the low-level features.
To mitigate the pose misalignment between the source and target persons, ~\cite{Siarohin2018} uses part-wise affine transformation with a modified feature fusion mechanism to warp the appearance features onto the target pose.
Later, extensive works have been presented to strengthen the modeling ability of body deformation and feature transfer with different methods, including 3D surface models~\cite{Neverova2018,Li2019,Grigorev2019}, local attention~\cite{Zhu2019,Ren2020} and optical flow~\cite{Wang2018video}.
\dif{
~\cite{Li2020self} and~\cite{Wang2021hierarchical} propose a rectification strategy in a self-learning way and hierarchical information framework, respectively, for human parsing, which benefits the downstream pose transfer task.
}
However, the warping methods commonly struggle with \emph{pose ambiguity} when \dif{the} viewpoint changes, occlusions occur, or even transferring a complicated pose in many situations.
To address the pose ambiguity, a series of works~\cite{Wang2018video,Liu2019} use predictive \dif{branches} to illuminate and replenish new contents for invisible regions.
When the hallucinated contents have a different context style than the local-warped ones, generated images will have a low visual fidelity due to \emph{appearance inconsistency}.
One of the main reasons for pose ambiguity and appearance inconsistency is that the commonly used reconstruction loss and the adversarial generative loss only constrain the synthesized image generation \dif{at the image level}.

Towards alleviating the mentioned limitations, it is important to disentangle the pose and appearance information, and exploit the disentangled pose and appearance feature consistencies between the synthesized and real images,
i.e., the synthesized target image should have a similar high-level appearance feature to the real target person as well as a similar high-level pose feature to the real source person.
The disentangled pose and appearance feature consistencies can constrain the training \dif{at the feature level} and lead to a more consistent and realistic synthesized result. 
\dif{
In CDMS~\cite{Zhou2022consistency}, a multi-mutual consistency learning strategy is proposed for the human pose segmentation task, showing the importance of feature consistency for distinguishing the human pose.
}

In this paper, we propose a pose transfer network with augmented \textbf{D}isentangled \textbf{F}eature \textbf{C}onsistency ({\ourmethod}) to facilitate human pose transfer.
{\ourmethod} contains a pose feature encoder and a static feature encoder to extract pose and appearance features from the source and target person, respectively.
In \dif{the} pose feature encoder, we integrated a pre-trained pose estimator such as OpenPose~\cite{Cao2019} to extract the keypoint heatmaps.
Notice that the pose estimator is pre-trained on COCO keypoint challenge dataset~\cite{Lin2014}, which is not any dataset deployed in our experiments.
As shown in Figure~\ref{Fig:sub1_heatmap_comparsion}, though the pre-trained pose estimator can predict pose heatmaps for unseen subjects in our dataset, it cannot generalize well and the heatmaps have much noise, which hinders subsequent pose transfer.
Further, in order to remedy the distortion of the extracted keypoints caused by the distribution shift from \dif{the} pose estimator, we introduce a keypoint amplifier to eliminate the noise in keypoint heatmaps.
An image generator synthesizes a realistic image of the target person conditioned on the disentangled pose and appearance features. 
The feature encoders and image generator empower {\ourmethod} to enable us to present novel feature-level pose and appearance consistency losses~\cite{Zhu2017}. 
These losses reinforce the consistency of pose and appearance information in the feature space and simultaneously maintain visual fidelity.
Additionally, to further improve the robustness and generality of {\ourmethod}, by disentangling the pose information from different source persons, we present a novel data augmentation scheme that builds an extra unpaired support dataset as the source images, which provides different persons with unseen poses in the training set and augmented consistency constraints.

% We also notice that the commonly used datasets and benchmarks~\cite{Zheng2015,Liu2016} only focus on the persons in the real world with a low-resolution, which is not sufficient to verify the effectiveness of each algorithm on other domains, such as computer animation~\cite{Aberman2019learning}.
\dif{
We also notice that the commonly used real-person datasets and benchmarks~\cite{Zheng2015,Liu2016} usually do not have the image of the target person with the desired pose from another source person, which \dif{is the ground truth}.
It is common practice to use a target person image directly from \dif{the} testing dataset to provide the pose information during the evaluation process.
Thus this practice raises the risk of leaking information and is also inconsistent with the usage in the real-world (i.e., the pose information is from \emph{another source person}).
In order to be consistent with the real-world application and to better evaluate the proposed method, 
inspired by~\cite{Aberman2019learning}, 
we collect an animation character image dataset named {\mixamodata} from Adobe Mixamo~\cite{Mixamo2018}, 
a 3D animation library, 
to accurately generate different characters performing identical poses as a benchmark to assess the human pose transfer between different people. 
}
% inspired by~\cite{Aberman2019learning}, 
% we collect an animation character image dataset named {\mixamodata} from Adobe Mixamo~\cite{Mixamo2018}, 
% a 3D animation library, 
% to accurately generate different characters performing the identical poses as a benchmark to assess the human pose transfer between different people. 
{\mixamodata} contains four different animation characters performing 15 kinds of poses. To further evaluate the {\ourmethod}, we also modify a real person dataset called {\edndata} upon~\cite{Chan2019}, which contains 10K high-resolution images for four real subjects performing different poses. 
The experimental results on these two datasets demonstrate that our model can effectively synthesize realistic images and conduct pose transfer for both the animation characters and real persons.

In summary, our contributions are as follows:
\begin{itemize}
	\item We propose a novel method {\ourmethod} for human pose transfer with two disentangled feature consistency losses to make the information between the real images and synthesized images consistent.
    \item We propose a novel data augmentation scheme that enforces augmented consistency constraints with an unpaired support dataset to further improve the generality of our model.
	\item We collect an animation character dataset {\mixamodata} as a new benchmark to enable the accurate evaluation of pose transfer between different people \dif{in the animation domain}.
	\item We conduct extensive experiments on datasets {\mixamodata} and {\edndata}, on which the empirical results demonstrate the effectiveness of our method.
\end{itemize}

\section{Related Work}
\label{sec:related}
% Generative adversarial networks (GANs)~\cite{Goodfellow2014} have achieved tremendous success in image-to-image translation tasks, whose goal is to convert images from one domain to a different domain.
\dif{
Generative adversarial networks~\cite{Goodfellow2014} and Diffusion models~\cite{Ho2020denoising} have achieved tremendous success in image generation tasks, whose goal is to generate high-fidelity images based on other images or text prompts from a different domain.
}
Pix2Pix~\cite{Isola2017} proposes a framework based on cGANs~\cite{Mirza2014} with an encoder-decoder architecture~\cite{Hinton2006};
CycleGAN~\cite{Zhu2017} addressed this problem by using cycle-consistent GANs; DualGAN~\cite{Yi2017} and~\cite{Hoshen2018} are also unsupervised image-to-image translation methods trained on unpaired datasets.
Similarly, ~\cite{Liu2017,Bousmalis2017,Huang2018} are also image-to-image translation techniques, but they try to generate a dataset of the target domain with labels for domain adaptation tasks.
The above works can be exploited as a general approach in the human pose transfer task, while the precondition is that they have a specific image domain \dif{that} can be converted to the synthesized image domain, e.g., using a pose estimator~\cite{Cao2017} to generate a paired skeleton image dataset.
\dif{
Based on the diffusion model, Diffustereo~\cite{Shao2022diffustereo} proposes a diffusion kernel and stereo constraints for 3D human reconstruction from sparse cameras.
MotionDiffuse~\cite{Zhang2022motiondiffuse} leverages the diffusion model on the text-driven motion generation task.
In this work, we focus on the 2D pose-guided motion transfer task, which differs from the above 3D reconstruction and test-driven tasks.
}
Different from the image-to-image translation methods, {\ourmethod} improved the quality of the synthesized image by adding consistency constraints in the feature space.

Recently, there \dif{have been} a growing number of human pose transfer methods with specifically designed modules.
% ~\cite{Lassner2017,Zhu2019,Li2019}.
One branch is the spatial transformation methods~\cite{Siarohin2018,Dong2018,Li2019}, aiming to build the deformation mapping of the keypoint correspondences in the human body.
By leveraging the spatial transformation capability of CNN, ~\cite{Jaderberg2015} presented the spatial transformer networks (STN) that approximate the global affine transformation to warp the features.
% ~\cite{Balakrishnan2018} introduced a warping method for image and corresponding pose by spatial transformer networks (STN)~\cite{Jaderberg2015}.
Following STN, several variant works~\cite{Zhang2017,Lin2017,Jiang2019} have been proposed to synthesize images with better performance. 
\dif{
~\cite{Wang2020paying} introduced an external eye-tracking dataset and two cascaded attention modules for comprehensive pose segmentation.
~\cite{Wang2021hierarchical} incorporated three different inference processes to detect each part of the human body.
}
~\cite{Balakrishnan2018} used image segmentation to decompose the problem into modular subtasks for each body part and then integrated all parts into the final result.  
~\cite{Siarohin2018} built deformable skip connections to move information and transfer textures for pose transfer.
Monkey-Net~\cite{Siarohin2019cvpr} encoded pose information via dense flow fields generated from keypoints learned in a self-supervised fashion.
First-Order Motion Model~\cite{Siarohin2019nips} decoupled appearance and pose and proposes to use learned keypoints and local affine transformations to generate image animation.
~\cite{Liu2019} integrated the human pose transfer, appearance transfer, and novel view synthesis into one unified framework by using SMPL~\cite{Loper2015} to generate a human body mesh. 
The spatial transformation methods usually implicitly assume that the warping operation can cover the whole body.
However, when \dif{the} viewpoint changes, and occlusions occur, the above assumption can not hold, leading to \emph{pose ambiguity} and performance dropping.

Another branch methods are pose-guided and aim to predict new appearance contents in uncovered regions to handle the pose ambiguity problem.
One of the earliest works, PG$^{2}$~\cite{Ma2017}, presented a two-stage method using U-Net to synthesize the target person with arbitrary poses.
~\cite{Ma2018} further decomposed the image into the foreground, background, and pose features to achieve more precise control of different information.
~\cite{Si2018} introduced a multi-stage GAN loss and synthesized each body part, respectively.
~\cite{Neverova2018} leveraged the DensePose~\cite{Alp2018} rather than the commonly used 2D key-points to perform accurate pose transfer.
~\cite{Chan2019} learned a direct mapping from the skeleton images to synthesized images with corresponding \dif{poses} based on the architecture of Pix2PixHD~\cite{Wang2018high}.
PATH~\cite{Zhu2019} introduced cascaded attention transfer blocks (PATBs) to refine pose and appearance features simultaneously.
Inspired by PATH, PMAN~\cite{Chen2021} proposed a progressive multi-attention framework with memory \dif{networks} to improve image quality.
However, some of these methods~\cite{Neverova2018,Zheng2019,Wang2018video} focused on synthesizing results at the image level (i.e., adversarial and reconstruction losses), thus leading to appearance inconsistency when predicted local contents are not consistent with the surrounding contexts.
\diff{Some works~\cite{Zhao2021real,Shen2021distilled} designed the light weighted networks to accelerate the training and inference process. 
Our method can also benefit from these light weighted networks to achieve high efficiency human pose transfer.
}

In contrast, our method learns to disentangle and reassemble the pose and appearance in the feature space.
One \dif{similar} work close to ours is C$^{2}$GAN~\cite{Tang2019} which consists of three generation cycles (i.e., one for image generation and two for keypoint generation).
C$^{2}$GAN explored the cross-modal information in the \dif{image level} at the cost of model complexity and training instability while {\ourmethod} only introduced two feature consistency losses into the full objective, which kept the model simple and effective. 
By disentangling the pose and appearance features, we can enforce the feature consistencies between the synthesized and real images and leverage the pose features from an unpaired dataset to improve performance.

\begin{figure*}[t]
    \centering
    \includegraphics[width=0.99\textwidth]{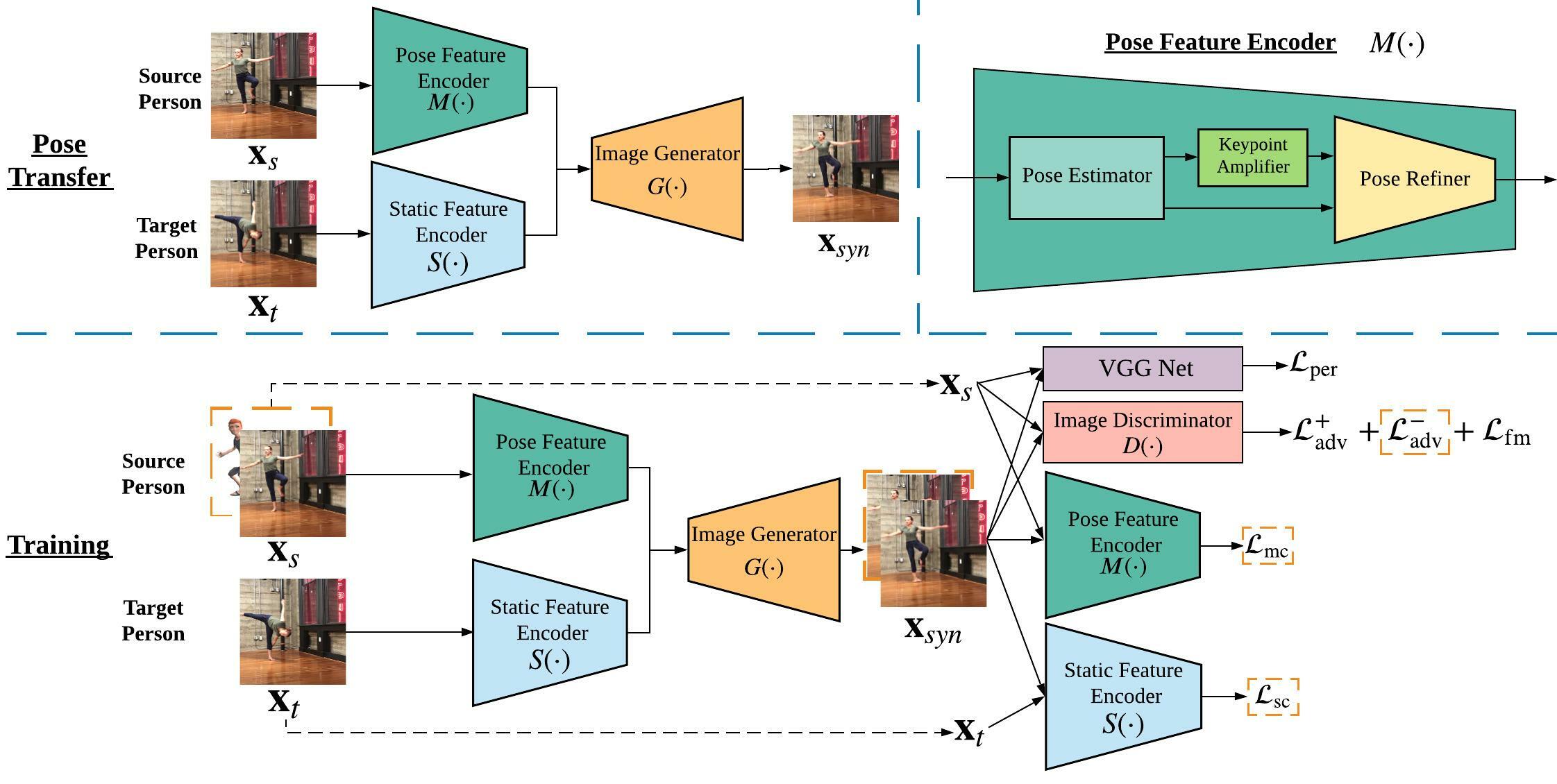}
    % \caption{The architecture of our proposed {\ourmethod}. Note that for training with support set only the losses $\mL_\radv^-, \mL_\rmc$ and $\mL_\rsc$ surrounded by orange dotted boxes are used. }
    \caption{
    \textbf{Upper left}: 
%    given a pair of images containing the source and target person,
    {\ourmethod} synthesizes an image of the target person performing the pose of the source person. 
    \textbf{Upper right}: the Pose Feature Encoder $M(\cdot)$ includes three components: pre-trained Pose Estimator, Keypoint Amplifier and Pose Refiner. 
    \textbf{Bottom}: the overview of training process of {\ourmethod}. Note that the images $x_s$ surrounded by orange dotted boxes is from the support set, and the augmented consistency losses $\mL_\rsup$ are sum of $\mL_\radv^-, \mL_\rmc$ and $\mL_\rsc$ surrounded by orange dotted boxes.}
    \label{Fig:Model}
%\vspace{-0.2cm}
\end{figure*}

\section{Methodology}
\label{sec:method}
\subsection{Overview}

The training and inference process of the proposed model is shown in Figure~\ref{Fig:Model}.
Given one image $\bx_{\rs}$ of a source person and another image $\bx_{\rt}$ of a target person, {\ourmethod} synthesizes an image $\bx_\rsyn$, which reserves a) the pose information, e.g., pose and location, of the source person in $\bx_{\rs}$, and b) the static information, e.g., person appearance and environment background, from the target image $\bx_{\rt}$.
For each image, {\ourmethod} attempts to disentangle the pose and static information into orthogonal features.
Specifically, {\ourmethod} consists of the following core components:
1) a \emph{\mfe} $M(\cdot)$, which extracts pose features $M(\bx)$ from an image $\bx$;
2) a \emph{\sfe} $S(\cdot)$, which extracts static features $S(\bx')$ from an image $\bx'$;
and 3) an \emph{\gen} $G(\cdot)$, which synthesizes an image $G(M(\bx), S(\bx'))$ based on the encoded pose and static features $M(\bx)$ and $S(\bx')$ from images $\bx$ and $\bx'$ separately.
In the remainder of this section, we describe the model architecture and introduce the training procedure, followed by the model instantiations.

\begin{figure*}[t]
    \centering
    \begin{subfigure}{.25\textwidth}
        \centering
        \includegraphics[width=0.95\linewidth]{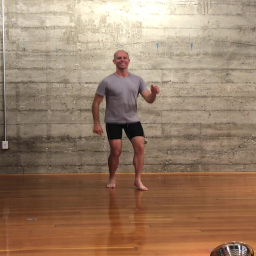}
        \label{Fig:subject1_frame007917}
        \caption{}
    \end{subfigure}%
    \begin{subfigure}{.25\textwidth}
        \centering
        \includegraphics[width=0.95\linewidth]{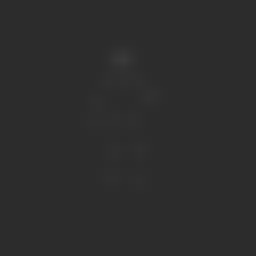}
        \label{Fig:subject1_frame007917_T1}
        \caption{}
    \end{subfigure}%
    \begin{subfigure}{.25\textwidth}
        \centering
        \includegraphics[width=0.95\linewidth]{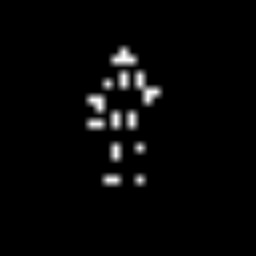}
        \label{Fig:subject1_frame007917_T0.1}
        \caption{}
    \end{subfigure}%
    \begin{subfigure}{.25\textwidth}
        \centering
        \includegraphics[width=0.95\linewidth]{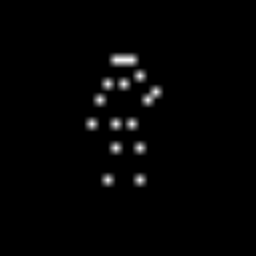}
        \label{Fig:subject1_frame007917_T0.01}
        \caption{}
    \end{subfigure}%
    \caption{Comparison of the amplified heatmaps of subject 1 in \edndata, which are generated from the Keypoint Amplifier with different temperature $T$. (a) the input image of subject 1. (b) the generated heatmaps with temperature $T=1$. (c) the generated heatmaps with temperature $T=0.1$. (d) the generated heatmaps with temperature $T=0.01$. We can observe that the heatmap with temperature $T=0.01$ is the most discriminative one, which is beneficial to later pose synthesis.} 
    \label{Fig:sub1_heatmap_comparsion}
%\vspace{-0.1cm}
% 1
\end{figure*}

%%%%%%%%%%%%%% Human Pose Image Generation %%%%%%%%%%%%%%%%%
\subsection{Pose Transfer Network Architecture}
\label{Section:Image Generation}
\subsubsection{\mfe}
Our designed {\mfe} consists of a {\pe} network, a {\ka} block, and a {\mfr} network.
% Even if the pre-trained pose estimator (OpenPose~\cite{Cao2019} in our implementation) is good enough, since it is pre-trained on COCO keypoint challenge dataset~\cite{Lin2014}, when applied on {\mixamodata} and {\edndata} datasets with different distributions, more noise are made on keypoint heatmaps.
Given a RGB image $\bx\in \R^{3\times H \times W}$ of height $H$ and width $W$, the pre-trained \textit{\pe}
aims at extracting pose information $P(\bx)$ from the image $\bx$.
Similar to~\cite{Cao2017}, the extracted pose information contains the downsampled keypoint heatmaps $\bh \in \R^{18 \times \frac{H}{8} \times \frac{W}{8}}$ and the part affinity fields $\bpaf\in \R^{38 \times \frac{H}{8} \times \frac{W}{8}}$.
The keypoint heatmaps $\bh$ store the heatmaps of 18 body parts, and the part affinity fields $\bpaf$ store the location and orientation for heatmaps of body parts and background, which has 38 ($=(18 + 1) \times 2$) channels.

As the pre-trained pose estimator (OpenPose~\cite{Cao2019} in our implementation) is pre-trained on COCO keypoint challenge dataset~\cite{Lin2014}, when applied on {\mixamodata} and {\edndata} datasets with different distributions, more noise \dif{is} made on keypoint heatmaps $\bh$.
% As there might be much noise in the estimated keypoint heatmaps $\bh$, 
To reduce the interference of noise, we apply a softmax function with a relatively small temperature $T$ (e.g., 0.01) as the \textit{\ka} to denoise the extracted keypoint heatmaps by increasing the gap between large and small values in the heatmaps and obtain the amplified heatmaps $\bh'$ by
\begin{align}
\bh' = \mathrm{softmax}\left(\frac{1}{T}\cdot \bh\right).
\end{align}

As shown in Figure~\ref{Fig:sub1_heatmap_comparsion}, by employing {\ka} on the input heatmaps, the small probability, e.g., 0.2, will be squeezed to almost 0.0. On the contrary, the large probability, e.g., 0.8, will be squeezed to almost 1.0.
Without the {\ka}, the generator may still synthesize blurry limbs for the low probability areas and twist the generated person.

Finally, the \textit{\mfr} takes both the part affinity fields $\bpaf$ and the amplified keypoint heatmaps $\bh'$ and produces the encoded pose feature vector $M(\bx)$. In this way, the pose information extracted from the {\pe} can be refined, and the influence caused by different limb ratios and/or camera angles and distances can be reduced.

\subsubsection{\sfe}
While the {\mfe} is not capable of extracting static information,
the static information, including background, personal appearance, etc., from another image $\bx'$, \dif{is} captured automatically by another module with the help of the full objective function.
Named as {\sfe}, this module extracts only static features $S(\bx')$ from $\bx'$.

\subsubsection{{\gen}}
Given pose features $M(\bx_\rs)$ extracted from a source image $\bx_\rs$ 
and static features $S(\bx_\rt)$ extracted from a target image $\bx_\rt$, the {\gen} outputs the synthesized image $\bx_{\rsyn}$ by
\begin{align}
    \label{Equation: syn_img}
    \bx_{\rsyn} = G\left(M(\bx_\rs), S(\bx_\rt)\right).
\end{align}

% It is noting that many existing methods (e.g., \cite{Chan2019}) attempt to learn pose-to-image or pose-to-appearance mapping solely via its generator, which is a difficult task on parameterizing the static information of the target in the generator. 
% Meanwhile, with the same generator, the desired pose information of $\bx_\rs$ may fail to synthesize if the desired pose is very different from the training poses.
\dif{
It is noted that many existing methods (e.g., \cite{Chan2019}) attempt to learn pose-to-image or pose-to-appearance mapping solely via its generator.
In that case, the generator has to learn three different functionalities: 1) memorizing the state information of the target person, 2) extracting representative pose features, and 3) combining the static and pose information to synthesize the target person image with the desired pose.
Even though the generator can memorize the state information $\bx_\rt$ of the target person perfectly, once the desired pose $\bx_\rs$ is very different from the poses in the training dataset (e.g., the distance from the camera, skeleton scale from different persons, and occlusions), it is too difficult for the generator to achieve the above second and third functionalities at the same time.
The results of Pix2Pix~\cite{Isola2017} and Everybody Dance Now (EDN)~\cite{Chan2019} in Section~\ref{sec:exp} also validate their disadvantages.
{\ourmethod}, instead, decomposes the above three functionalities into three network modules, including the pose feature encoder, static feature encoder, and image generator, and thus enables the reconstruction's quality improvement.
}
% {\ourmethod}, instead, extracts and utilizes decomposed static and pose features, which enables the reconstruction's quality improvement.

\subsection{Training {\ourmethod}}
We train the pose transfer network in an adversarial learning way with disentangled feature consistency losses as well as other objectives.
Basically, the model is trained with a set of images of the same person, possibly from one or several video clips.
To further improve the generalization ability of {\ourmethod}, we propose to train {\ourmethod} with a support set and the augmented consistency losses.
We show the ablation study results in Section~\ref{sec:abl_study}.

\subsubsection{Adversarial Training}
\label{Section:Adversarial Training}
We employ an \textit{\dis} (D) in an adversarial learning way to ensure the synthesized image $\bx_\rsyn$ borrows the pose and static information from the source and target images ($\bx_\rs$ and $\bx_\rt$) separately.
As both the source and target images during training contain the same person with the same appearance and background, they share almost the same static features, i.e., $S(\bx_\rt) \simeq S(\bx_\rs)$.
Therefore, the output of the model $\bx_{\rsyn}$ can also be treated as a reconstruction of the source image $\bx_\rs$, as the synthesized images contain the same pose features $M(\bx_\rs)$ as the source image.
This inspires us to resort to the conditional generative adversarial network (cGAN)~\cite{Isola2017}, where the {\dis} attempts to discern between the real sample $\bx_\rs$ and the generated image $\bx_{\rsyn}$, conditioned on the pose features $M(\bx_\rs)$ extracted from the source image. That is, the {\dis} attempts to fit $D (\bx_\rs, M(\bx_\rs))=1$ and $ D (\bx_\rsyn, M(\bx_\rs))) = 0$.
The adversarial loss is described as follows:
\begin{align}
\label{Equation: adv}
\mL_{\radv} = - (\mL_{\radv}^+ + \mL_{\radv}^-),
\end{align}
where
\begin{align}
\mL_{\radv}^+ &= \log D (\bx_\rs, M(\bx_\rs)),  \\
\mL_{\radv}^- &= \log \left(1 - D \left(\bx_\rsyn, M(\bx_\rs)\right)\right).
\end{align}

We enhance the {\dis} with a multi-scale discriminator $D = (D_1, D_2)$~\cite{Wang2018high} and include the discriminator feature matching loss $\mL_\rfm$ in our objective.
The feature matching loss is a weighted sum of feature losses from 5 different layers of the Image Discriminator, calculated by $L_1$ distance between the corresponding features of $\bx_\rs$ and $\bx_\rsyn$.

In order to increase the training stability and improve the synthesized image quality, we also add the perceptual loss $\mL_\rper$ ~\cite{Johnson2016} based on a pre-trained VGG network~\cite{Simonyan2014}.

\subsubsection{Disentangled Feature Consistency Losses}
The above adversarial training losses aim at penalizing discrepancy between the synthesized and source images directly in the raw image space.
To improve the accuracy and robustness of the pose transfer results, we also introduce two disentangled feature consistency losses in terms of \emph{pose} and \emph{static} features to \dif{ensure} the synthesized person looks like the target person and behaves as the source person separately.
The pose consistency loss $\mL_\rmc$ measures the differences between the synthesized and source images in the pose feature space, and the static consistency loss $\mL_\rsc$ measures the differences between the synthesized and target images in the static feature space.
They are both $L_1$ distances between the outputs from the corresponding encoders, formally defined as
\begin{align}
\mL_\rmc & = {\left\| M(\bx_\rsyn) - M(\bx_\rs) \right\|}_1, \\
\mL_\rsc & = {\left\| S(\bx_\rsyn) - S(\bx_\rt) \right\|}_1.
\end{align}

% \subsubsection{Training with Support Set}
\subsubsection{Augmented Consistency Loss}
\label{Section:Pose Support Set}
Through disentangling the pose feature from the source images, we find that images with different persons can also be passed into the training process as the source images $\bx_\rs$ to improve the generalization ability of our model.
Hence, we introduce a novel data augmentation method that extends the training dataset with the images of different persons, referred to as the \textit{support set}, providing many kinds of unseen poses.
Note that the subjects in \textit{support set} can be arbitrary and are different from the primary training dataset, so the ground-truth images with the target person performing the pose of the source person are not available at all.
As a result, the corresponding losses $\mL_\radv^+, \mL_\rper$ and $\mL_\rfm$ for the support set are not applicable,
and we optimize relevant objective terms, which \dif{are} defined by

\begin{align}
    \mL_\rsup = \lambda_\radv \mL_\radv^- + \lambda_\rmc \mL_\rmc + \lambda_\rsc \mL_\rsc
\end{align}
where the weights $\lambda_\radv, \lambda_\rmc, \lambda_\rsc$ are the weights for each loss.

\subsubsection{Full Objective}

By bringing all the objective terms together, we train all components jointly except for the {\pe} to minimize the full objective $\mL_\rfull$ below.
\begin{align}
\label{Equatio: full_loss}
    \mL_\rfull = & \lambda_\radv \mL_\radv + \lambda_\rfm \mL_\rfm + \lambda_\rper \mL_\rper \\ \nonumber
    & + \lambda_\rmc \mL_\rmc + \lambda_\rsc \mL_\rsc + \mL_\rsup
\end{align}
where $\lambda_\radv, \lambda_\rfm, \lambda_\rper$ are set to 1, 10, 10 following Pix2Pix~\cite{Isola2017} and Everybody Dance Now (EDN)~\cite{Chan2019},
while $\lambda_\rmc, \lambda_\rsc$ are set to 0.1, 0.01 by grid search. 
We set the $\lambda_\rsc$ to 0.01 comparing to the $\lambda_\rmc$ to balance the $\mL_\rmc$ and $\mL_\rsc$.

\subsection{Training and Inference Process}
\label{Section:Training and Inference Process}
For each subject in the training dataset (i.e., {\mixamodata} and {\edndata} in our experiments), we train one separate model following the same scheme of~\cite{Chan2019} (e.g., we trained four models for four subjects in EDN-10k dataset.)  
For the sake of comparison fairness, we also train all the baseline methods following the same scheme.

During the training stage, given a training dataset consisting of N images of one subject and a support set, for each training iteration, we randomly choose a pair of images as $x_s$ and $x_t$ from the training dataset and an image as $x_s$ from the support set respectively, pass them into the DFC-Net and train it using the full objective in Equation~\ref{Equatio: full_loss}.

During the inference stage, given a desired pose image $x_s$, we randomly choose an image $x_t$ from the training dataset and synthesize the result.
\dif{
For {\edndata} dataset, the pose image $x_s$ is chosen from the testing dataset with an unseen pose in the training process.
Even though the pose image $x_s$ and the target person image $x_t$ contain the same person (i.e., the ground truth of the pose image $x_s$ with another person is unavailable for real-world data), by passing through the pose feature encoder, the static information in the pose image $x_s$ is discarded, and only the keypoints information are preserved. 
For {\mixamodata} dataset, the pose image $x_s$ is chosen from the testing dataset, including the different person from the target person (e.g., the target person image $x_t$ is from Liam and the source person image $x_s$ is from Remy).
For both benchmark, {\ourmethod} has to extract the static features from the target person image $x_t$ and combines them with the pose features of the pose image $x_s$ to synthesize the final images where the poses are unseen during the training. Thus there is no information leakage.
}

\subsection{Implementation details}
We employed the pre-trained VGG-19~\cite{simonyan2014very} network part from~\cite{Cao2017} for the {\pe},
and adopted the similar approach in~\cite{Wang2018video} to build our network, the detailed designs are as follows:
\begin{itemize}
	\item \textbf{{\mfr}}: It is composed of a convolutional block, a channel-wise upsampling module, and five residual blocks~\cite{He2016}.
	Firstly, the convolutional block consists of a reflection padding layer, a $7 \times 7$ convolutional layer, a batch normalization layer, and ReLU. 
	The channel-wise upsampling module, which increases the number of \dif{channels} from 64 to 512, contains three convolutional blocks. Each block contains a $3 \times 3$ convolutional layer, a batch normalization layer, and ReLU.
	Each of the five residual blocks consists of two small convolutional blocks, and each block has a reflection padding layer, a $3 \times 3$ convolutional layer, \dif{and} a batch normalization layer. The first small convolutional block also has a ReLU at the end.
	\item \textbf{{\sfe}}: It firstly has the same convolutional block as in the Pose Refiner. 
	Then it contains three convolutional downsampling blocks, and each block consists of a $3 \times 3$ convolutional layer, a batch normalization layer, and ReLU.
	There are also five residual blocks following the downsampling blocks as the same as in the Pose Refiner.
	\item \textbf{{\gen}}: It is composed of four residual blocks, an upsampling module, and a convolutional block.
	Each of the residual blocks is the same as in the Pose Refiner and Static Feature Encoder.
	The upsampling module consists of three transposed convolutional blocks, and each block is composed of a $3 \times 3$ transposed convolutional layer, a batch normalization layer, and ReLU.
	The last convolutional block contains two reflection padding layers, two $7 \times 7$ convolutional layers, and a tangent function.
	\item \textbf{{\dis}}: It contains two discriminators at different scales, which are similar to \cite{Wang2018high}.
	\dif{Each discriminator} is composed of five convolutional blocks. 
	The first block has a $4 \times 4$ convolutional layer and LeakyReLU. 
	Each of the next three blocks has a $4 \times 4$ convolutional layer, a batch normalization layer, and Leaky ReLU.
	The last block only has a $4 \times 4$ convolutional layer.
\end{itemize}

\section{Experiments}
\label{sec:exp}

\subsection{Experimental Setup}
\label{Sec:Setup}
\subsubsection{Datasets}
We built {\supdata} as \dif{a} support set for data augmentation to boost the generality of {\ourmethod} and organized two datasets, {\mixamodata} and {\edndata}, to verify the effectiveness of the proposed {\ourmethod} for human pose transfer.

\begin{itemize}
	\item \textbf{\edndata}:
	We processed and tailored the dataset released by \cite{Chan2019} to build the {\edndata} dataset.
    The original dataset consists of five long target videos, each lasting from 8 minutes to 17 minutes, and split into the training and test set.
    We chose the first four subjects since subject five only performed less complex dance poses.
    In each video of the original dataset, a different subject performed a series of different motions, and the camera was fixed to keep the background unchanged.
    We chose four subjects from the original dataset, uniformly sampled 10k frames as the training set and 1k frames as the test set for each subject, 
    Since the original images have a large resolution of $1024 \times 512$ and most areas are the fixed background,
    we cropped all frames to the middle $512 \times 512$ square areas and resized them to $256 \times 256$.
    
	\item \textbf{\mixamodata}: We randomly chose 4 characters, \textit{Andromeda}, \textit{Liam}, \textit{Remy}, and \textit{Stefani}, with 30 different pose sequences from Mixamo. 
	To render the 3D animations into 2D images, we loaded each character performing each pose sequence on a white background into Blender~\cite{Blender2018}, placed two cameras in front of and behind the character, and took the images.
	We centered the characters in the images according to their keypoints and resized them to 256$\times$256.
	{\mixamodata} were \dif{split} into training and test sets. For each character, the training set contains 1488 images with 15 poses, and the test set contains 1185 images with 15 other poses.

    \item \textbf{\supdata}: For data augmentation, we built a \textit{support set} by rendering 15684 images of six new characters from Mixamo~\cite{Mixamo2018} with another 15 unseen poses in the same way as {\mixamodata}.
    Since it is unnecessary to contain the same person as the target person image, {\ourmethod} leveraged the support set as the source person images $x_s$. 
    When training on both {\edndata} and {\mixamodata}, we use {\supdata} as the support set.
    Note that {\supdata} has a totally different distribution from {\edndata} but still gains a huge improvement shown in Section~\ref{sec:ablation}.
\end{itemize}

% \begin{table}[t]
% \centering
% \resizebox{0.7\columnwidth}{!}
% {\begin{tabular}{l|rrr}
% \toprule
% Method & MSE($\downarrow$) & PSNR($\uparrow$)  & SSIM($\uparrow$)  \\ \midrule
% NN & 26.3576 & 34.3209 & 0.7380 \\
% CycleGAN & 27.1219 & 33.8762 & 0.7372 \\
% Pix2Pix & 23.2877 & 34.5397 & 0.7935 \\
% EDN & 26.6151 & 34.5217 & 0.7933 \\
% LWG & 23.0527 & 34.5972 & 0.7817 \\
% MKN & 30.8873 & 33.6933 & 0.7308 \\
% FOMM & 28.6544 & 33.8486 & 0.7392 \\
% \cmidrule(lr){1-4}
% \textbf{Ours} & \textbf{22.2304} & \textbf{34.7565} & \textbf{0.7973} \\ \bottomrule
% \end{tabular}}
% \caption{Comparisons of average results on {\mixamodata} with best results in bold.}
% \label{tab:ave-results-mixamo}
% %\vspace{-0.4cm}
% \end{table}

Note that the experiments on {\edndata} only include the pose transfer on the same person because the \dif{ground truth} of different people carrying the same pose are unavailable.
For {\mixamodata}, since we can manipulate different characters to do the same action, the experiments include the pose transfer both on the different people, e.g., transfer the unseen pose of \textit{Liam} to \textit{Andromeda}, and the same person, e.g., transfer the unseen pose of \textit{Andromeda} to herself.
% The experiments on {\edndata} only include the pose transfer on the same person because the ground-truth of different people carrying the same pose are unavailable.

% \paragraph{Baseline methods}
\subsubsection{Baseline Methods}
We compared our {\ourmethod} with the following competitive baselines:
\begin{itemize}
	\item \textbf{Nearest Neighbors (NN)}: For each source person image $\bx_\rs$, we chose the image $\bx'$ in the training set $ \mathcal{D}_{\rtrain}$ with the lowest mean square error (MSE) between the pose information $P(\bx_\rs)$ and $P(\bx')$ as $\bx_\rsyn$.
	\begin{align}
        \bx_\rsyn & = {\arg\min}_{\bx' \in \mathcal{D}_{\rtrain}}{\left\| P(\bx_\rs) - P(\bx') \right\|}^2_2.
    \end{align}
    The pose information was extracted by the same {\pe} as in our method.

    \item \textbf{Pose-guided Methods}: We chose CycleGAN~\cite{Zhu2017}, Pix2Pix~\cite{Isola2017} and Everybody Dance Now (EDN)~\cite{Chan2019} as the baselines. 
    They all took the skeleton images as input instead of the original images, \dif{and we} employed a pre-trained pose estimator~\cite{Cao2017} to extract keypoints, used OpenCV~\cite{opencv2000} to connect pairs of keypoints with different colors to generate the skeleton images.
    To ensure fair comparisons, the face GAN and face keypoint estimator in EDN were not adopted in our implementation, as they are independent components and can be seamlessly adopted by other learning-based baselines. 
    \item \textbf{Spatial Transformation Methods}: We selected Liquid Warping GAN (LWG)~\cite{Liu2019}, Monkey-Net (MKN)~\cite{Siarohin2019cvpr} and First Order Motion Model (FOMM)~\cite{Siarohin2019nips}.
    LWG calculates the flow fields with additional 3D human models and integrates the human pose transfer, appearance transfer, and novel view synthesis into one unified framework.
    MKN and FOMM are both object-agnostic frameworks using learned keypoints to generate image animation in a self-supervised fashion.
\end{itemize}

% \subsubsection{Tasks Settings}
% \input{exp-setting.tex}

% \begin{table}[t]
% \centering
%  \resizebox{0.7\columnwidth}{!}
% {\begin{tabular}{l|rrrr|r}
% \toprule
% Method & \textit{Subject1} & \textit{Subject2} & \textit{Subject3} & \textit{Subject4} & Average \\ \midrule
% NN & 30.7957 & 32.6439 & 31.3307 & 34.2481 & 32.2546 \\
% CycleGAN~\cite{Zhu2017} & 30.0577 & 29.2421 & 29.7007 & 31.3398 & 30.0851 \\
% Pix2Pix~\cite{Isola2017} & 30.6020 & 32.0304 & 30.5843 & 34.6860 & 31.9757 \\
% EDN~\cite{Chan2019} & 30.7109 & 32.8639 & 30.9957 & 35.0369 & 32.4019 \\
% LWG~\cite{Liu2019} & 31.0085 & 31.7812 & 31.0503 & 35.0118 & 32.2129 \\
% MKN~\cite{Siarohin2019cvpr} & 31.3094 & 33.2530 & 30.6493 & 34.9108 & 32.5306 \\
% FOMM~\cite{Siarohin2019nips} & 31.5272 & 33.3145 & 31.3099 & 34.9612 & 32.7782 \\
% \cmidrule(lr){1-6}
% \textbf{Ours} & \textbf{31.5978} & \textbf{33.3509} & \textbf{31.4159} & \textbf{35.0718} & \textbf{32.8591} \\ \bottomrule
% \end{tabular}}
% \caption{Comparisons on {\edndata} in terms of PSNR with the best results (highest values) in bold.}
% \label{tab:psnr-results-edn}
% \end{table}

% \paragraph{Evaluation metrics}
\subsubsection{Evaluation Metrics}
We evaluated the quality of the synthesized images with three commonly used metrics:
\begin{itemize}
	\item \textbf{MSE}: The mean squared error between the values of pixels of synthesized images and ground-truth images.
	\textit{Lower} MSE values are better.
	\item \textbf{PSNR}: The peak signal-to-noise ratio, which provides an empirical measure of the quality of synthesized images regarding ground-truth images.
	\textit{Higher} PSNR values are better.
	\item \textbf{SSIM}: Structural similarity~\cite{Wang2004}, which is another perceptual metric that quantifies the quality of synthesized images given ground-truth images and focuses more on structural information (e.g., light).
	\textit{Higher} SSIM values are better.
        \dif{
        \item \textbf{IS}: Inception Score~\cite{Salimans2016improved} is a metric for estimating the quality of the synthetic images based on the Inception-V3 model~\cite{Szegedy2016rethinking}. 
        \textit{Higher} IS values are better.
        \item \textbf{FID}: Frechet Inception Distance~\cite{Heusel2017gans} is also an Inception-V3-based metric to evaluate the synthetic images according to the statistics of the synthetic images.
        \textit{Lower} FID values are better.
        }
\end{itemize}

We calculated the average scores among all pairs of synthesized and ground-truth images on the test set. On {\mixamodata}, for each character as the target person, we reported the average metrics of 4 different characters as the source person. While on {\edndata}, we reported metrics of the task on the same person for every subject.

% \begin{table}[t]
% \centering
% \resizebox{0.7\columnwidth}{!}
% {\begin{tabular}{l|rrr}
% \toprule
% Method & MSE($\downarrow$) & PSNR($\uparrow$)  & SSIM($\uparrow$)  \\ \midrule
% NN & 43.1572 & 32.2546 & 0.7611 \\
% CycleGAN & 66.1287 & 30.0851 & 0.5964 \\
% Pix2Pix & 46.9074 & 31.9757 & 0.7658 \\
% EDN & 42.5696 & 32.4019 & 0.7956 \\
% LWG & 42.4368 & 32.2129 & 0.7709 \\
% MKN & 37.1098 & 32.5306 & 0.8111 \\
% FOMM & 37.3648 & 32.7782 & 0.7908 \\
% \cmidrule(lr){1-4}
% \textbf{Ours} & \textbf{36.2377} & \textbf{32.8591} & \textbf{0.8269} \\ \bottomrule
% \end{tabular}}
% \caption{Comparisons of average results on {\edndata} with best results in bold.}
% \label{tab:ave-results-edn}
% %\vspace{-0.4cm}
% \end{table}

\begin{table}[t]
    \begin{minipage}[t]{.48\linewidth}
    \centering
    \caption{Comparisons on {\edndata} in terms of MSE with the best results (lowest values) in bold.}
    \label{tab:mse-results-edn}
    \resizebox{\textwidth}{!}{
        \begin{tabular}{l|rrrr}
        \toprule
        Method & \textit{Subject1} & \textit{Subject2} & \textit{Subject3} & \textit{Subject4} \\ \midrule
        NN & 54.9448 & 36.2267 & 55.2041 & 26.2531  \\
        \dif{CycleGAN~\cite{Zhu2017}} & 64.1959 & 77.4336 & 70.3681 & 52.5171  \\
        \dif{Pix2Pix~\cite{Isola2017}} & 58.3633 & 43.0771 & 62.0203 & 24.1688  \\
        \dif{EDN~\cite{Chan2019}} & 56.3549 & 36.3887 & 55.9625 & 21.5724  \\
        \dif{LWG~\cite{Liu2019}} & 51.6246 & 43.2884 & 53.4031 & 21.4314  \\
        \dif{MKN~\cite{Siarohin2019cvpr}} & 48.1603 & 30.9902 & \textbf{47.6255} & 21.6634  \\
        \dif{FOMM~\cite{Siarohin2019nips}} & 46.2852 & 30.7603 & 51.1431 & 21.2709  \\
        \cmidrule(lr){1-5}
        \textbf{Ours} & \textbf{45.2043} & \textbf{30.4782} & 48.5436 & \textbf{20.7248} \\ \bottomrule
        \end{tabular}
    }
    \end{minipage}
    \begin{minipage}[t]{.01\linewidth}
        \hfill
    \end{minipage}
    \hfill
    \begin{minipage}[t]{.48\linewidth}
    \centering
    \caption{Comparisons on {\edndata} in terms of SSIM with the best results (highest values) in bold.}
    \label{tab:ssim-results-edn}
    \resizebox{\textwidth}{!}
    {
        \begin{tabular}{l|rrrr}
        \toprule
        Method & \textit{Subject1} & \textit{Subject2} & \textit{Subject3} & \textit{Subject4} \\ \midrule
        NN & 0.6138 & 0.8253 & 0.7616 & 0.8437 \\
        \dif{CycleGAN~\cite{Zhu2017}} & 0.5256 & 0.4911 & 0.5869 & 0.7821  \\
        \dif{Pix2Pix~\cite{Isola2017}} & 0.6238 & 0.8040 & 0.7585 & 0.8767  \\
        \dif{EDN~\cite{Chan2019}} & 0.6205 & 0.8445 & 0.8233 & 0.8939  \\
        \dif{LWG~\cite{Liu2019}} & 0.6394 & 0.8375 & 0.7434 & 0.8634  \\
        \dif{MKN~\cite{Siarohin2019cvpr}} & 0.7007 & 0.8503 & 0.8030 & 0.8904  \\
        \dif{FOMM~\cite{Siarohin2019nips}} & 0.6645 & 0.8445 & 0.7896 & 0.8649  \\
        \cmidrule(lr){1-5}
        \textbf{Ours} & \textbf{0.7083} & \textbf{0.8670} & \textbf{0.8241} & \textbf{0.9083} \\ \bottomrule
        \end{tabular}
    }
    \end{minipage}
\end{table}

\begin{table}[t]
  \begin{minipage}[t]{.48\linewidth}
    \centering
    \caption{Comparisons on {\edndata} in terms of PSNR with the best results (highest values) in bold.}
    \label{tab:psnr-results-edn}
    \resizebox{\textwidth}{!}
    {
        \begin{tabular}{l|rrrr}
        \toprule
        Method & \textit{Subject1} & \textit{Subject2} & \textit{Subject3} & \textit{Subject4} \\ \midrule
        NN & 30.7957 & 32.6439 & 31.3307 & 34.2481  \\
        \dif{CycleGAN~\cite{Zhu2017}} & 30.0577 & 29.2421 & 29.7007 & 31.3398  \\
        \dif{Pix2Pix~\cite{Isola2017}} & 30.6020 & 32.0304 & 30.5843 & 34.6860  \\
        \dif{EDN~\cite{Chan2019}} & 30.7109 & 32.8639 & 30.9957 & 35.0369  \\
        \dif{LWG~\cite{Liu2019}} & 31.0085 & 31.7812 & 31.0503 & 35.0118  \\
        \dif{MKN~\cite{Siarohin2019cvpr}} & 31.3094 & 33.2530 & 30.6493 & 34.9108  \\
        \dif{FOMM~\cite{Siarohin2019nips}} & 31.5272 & 33.3145 & 31.3099 & 34.9612  \\
        \cmidrule(lr){1-5}
        \textbf{Ours} & \textbf{31.5978} & \textbf{33.3509} & \textbf{31.4159} & \textbf{35.0718} \\ \bottomrule
        \end{tabular}
    }
    \end{minipage}
    \begin{minipage}[t]{.01\linewidth}
        \hfill
    \end{minipage}
        \hfill
    \begin{minipage}[t]{.48\linewidth}
    \centering
    \caption{\dif{Comparisons on {\edndata} in terms of IS with the best results (highest values) in bold.}}
    \label{tab:IS-results-edn}
    \resizebox{\textwidth}{!}
    {
        \begin{tabular}{l|rrrr}
        \toprule
        Method & \textit{Subject1} & \textit{Subject2} & \textit{Subject3} & \textit{Subject4} \\ \midrule
        NN & 3.1284 & 3.3052 & 3.1525 & 3.3271 \\
        CycleGAN~\cite{Zhu2017} & 2.9294 & 2.8902 & 2.9776 & 3.0343 \\
        Pix2Pix~\cite{Isola2017} & 3.1903 & 3.2966 & 3.1864 & 3.5012  \\
        EDN~\cite{Chan2019} & 3.1802 & 3.4328 & 3.4083 & 3.5316 \\
        LWG~\cite{Liu2019} & 3.1774 & 3.4035 & 3.1365 & 3.4122  \\
        MKN~\cite{Siarohin2019cvpr} & 3.3481 & 3.4680 & 3.3793 & 3.5238 \\
        FOMM~\cite{Siarohin2019nips} & 3.2794 & 3.4075 & 3.3281 & 3.4019 \\
        \cmidrule(lr){1-5}
        \textbf{Ours} & \textbf{3.3502} & \textbf{3.5227} & \textbf{3.4240} & \textbf{3.5682}  \\ \bottomrule
        \end{tabular}
    }
  \end{minipage}
\end{table} 

\subsection{Quantitative Evaluations}
\label{Sec:Results}
\subsubsection{Results on {\edndata}}
% Tables~\ref{tab:mse-results-edn},~\ref{tab:ssim-results-edn} and~\ref{tab:psnr-results-edn} depict results in terms of MSE, SSIM and PSNR on the {\edndata} dataset.
% Table~\ref{tab:ave-results-edn} depicts the average results of 4 subjects on the {\edndata} dataset.
\dif{
Tables~\ref{tab:mse-results-edn},~\ref{tab:ssim-results-edn}, ~\ref{tab:psnr-results-edn},~\ref{tab:IS-results-edn}, and~\ref{tab:FID-results-edn} depict results in terms of MSE, SSIM, PSNR, IS, and FID on the {\edndata} dataset.
Table~\ref{tab:ave-results-edn} provides comparisons on {\edndata} in terms of average results of the above five metrics over four subjects.
}
The experimental results validated the advances of {\ourmethod} for real images:

\begin{itemize}
	\item Our method consistently outperformed all the baseline methods on all subjects.
	When synthesizing real person images, the most significant result is that on \textit{Subject1}, our method achieved 45.2043 MSE while NN, CycleGAN, Pix2Pix, EDN, LWG, and MKN only got MSE values of 54.9448, 64.1959, 58.3633, 56.3549, 51.6246 and 48.1603 in Table~\ref{tab:mse-results-edn},
    which indicates the images generated by our method have clear details.
    % As shown in Table~\ref{tab:ssim-results-edn}, 
    \dif{
    From Tables~\ref{tab:IS-results-edn} and~\ref{tab:FID-results-edn}, {\ourmethod} also excelled other baselines for all four subjects according to the Inception Score (IS) and Frechet Inception Distance (FID).
    As shown in Table~\ref{tab:ave-results-edn}, 
    }
    our method also achieved the highest average SSIM of 0.8269 for all subjects, while no other methods except MKN got SSIM score greater than 0.8, which shows that our synthesized images are more realistic and suitable for the human visual system.
% ##########################
    
    \item Secondly, we noticed CycleGAN has the worst results, which are 64.1959, 77.4336, and 70.3681 for MSE scores, and 0.5256, 0.4911, and 0.5869 for SSIM scores on \textit{Subject1}, \textit{Subject2}, and \textit{Subject3} in the Tables~\ref{tab:mse-results-edn} and ~\ref{tab:ssim-results-edn} respectively.
    We argue that CycleGAN is better at transferring the color or style for images from two domains rather than \dif{changing} the geometry of the images, such as \dif{recovering} the human appearance from the human skeleton because CycleGAN aims to learn a mapping from unpaired images directly.
    This property of CycleGAN is also supported by its inferior PSNR scores \dif{compared} with other methods.
    
    \item We could observe that NN can achieve low scores of MSE in Table~\ref{tab:mse-results-edn}. 
    Since there are a lot of training images, it is easier for NN to find an image whose motion is very close to the desirable motion.
	Moreover, the fixed background also makes NN have higher scores of SSIM and PSNR in Tables~\ref{tab:ssim-results-edn} and ~\ref{tab:psnr-results-edn}, while other methods have to learn to generate an accurate background.
	But the images generated by NN usually do not perform the desired motions,
	and do not have any temporal coherence in motion when the input is a motion sequence since the results only depend on the training set.
    
%     \item EDN and MKN also provided good results, especially on SSIM metric, e.g., 0.8939 and 0.8904 for \textit{Subject4} in Table~\ref{tab:ssim-results-edn}.
% 	Pix2Pix and EDN also provide good results, especially on SSIM metric, e.g., 0.8767 and 0.8939 for \textit{Subject4} in Table~\ref{tab:ssim-results-edn}.
% 	Taking \textit{Subject4} as an example, Pix2Pix and EDN provide the SSIM of 0.8767 and 0.8939 in Table~\ref{tab:ssim-results-edn} which are higher than the result of NN.
% 	The higher SSIM values show that Pix2Pix and EDN can synthesize images closer to the ground-truths.
% 	even though the value of each pixel may be different, which causes higher MSE.

	\item Moreover, we observed that EDN, LWG, MKN, and FOMM also provided good results, especially on MSE metric, e.g., 42.5696, 42.4368, 37.1098, and 37.3648 average values for all subjects in Table~\ref{tab:mse-results-edn}, comparing with other baselines.
	Taking \textit{Subject2} as an example, LWG, MKN, and FOMM provided the SSIM of 0.8445, 0.8375, 0.8503, and 0.8445 in Table~\ref{tab:ssim-results-edn} which are higher than the results of NN, CycleGAN, and Pix2Pix.
	The higher SSIM values show that LWG, MKN, and FOMM can synthesize images closer to the ground truths.
% #########################
\end{itemize}

\begin{table}[t]
  \begin{minipage}[t]{.48\linewidth}
    \centering
    \caption{\dif{Comparisons on {\edndata} in terms of FID with the best results (lowest values) in bold.}}
    \label{tab:FID-results-edn}
    \resizebox{0.88\textwidth}{!}
    {
        \begin{tabular}{l|rrrr}
        \toprule
        Method & \textit{Subject1} & \textit{Subject2} & \textit{Subject3} & \textit{Subject4} \\ \midrule
        NN & 24.8053 & 19.2783 & 22.4787 & 23.3219  \\
        CycleGAN~\cite{Zhu2017} & 37.8460 & 38.8926 & 35.3055 & 32.9549  \\
        Pix2Pix~\cite{Isola2017} & 23.5316 & 21.3092 & 24.7829 & 17.5752  \\
        EDN~\cite{Chan2019} & 23.8172 & 18.3454 & 19.0175 & 14.4926  \\
        LWG~\cite{Liu2019} & 22.3348 & 18.1062 & 25.7232 & 19.7245  \\
        MKN~\cite{Siarohin2019cvpr} & 19.4209 & 16.5735 & 19.5568 &  14.4617 \\
        FOMM~\cite{Siarohin2019nips} & 20.3571 & 17.6391 & 21.3258 & 18.7283  \\
        \cmidrule(lr){1-5}
        \textbf{Ours} & \textbf{18.3029} & \textbf{15.2877} & \textbf{17.8782} & \textbf{13.7508}  \\ \bottomrule
        \end{tabular}
    }
    \end{minipage}
    \begin{minipage}[t]{.01\linewidth}
        \hfill
    \end{minipage}
        \hfill
    \begin{minipage}[t]{.48\linewidth}
    \centering
    \caption{\dif{Comparisons on {\edndata} in terms of the average results of the 5 metrics over 4 subjects.}}
    \label{tab:ave-results-edn}
    \resizebox{\textwidth}{!}
    {
        \begin{tabular}{l|rrrrr}
        \toprule
        Method & MSE($\downarrow$) & SSIM($\uparrow$)  & PSNR($\uparrow$) & IS($\uparrow$) & FID($\downarrow$) \\ \midrule
        NN & 43.1572 & 0.7611 & 32.2546 & 3.2283 & 22.4711 \\
        CycleGAN~\cite{Zhu2017} & 66.1287 & 0.5964 & 30.0851 & 2.9579 & 36.2498 \\
        Pix2Pix~\cite{Isola2017} & 46.9074 & 0.7658 & 31.9757 & 3.2936 & 21.7997 \\
        EDN~\cite{Chan2019} & 42.5696 & 0.7956 & 32.4019 & 3.3882 & 18.9182 \\
        LWG~\cite{Liu2019} & 42.4368 & 0.7709 & 32.2129 & 3.2824 & 21.4496 \\
        MKN~\cite{Siarohin2019cvpr} & 37.1098 & 0.8111 & 32.5306 & 3.4298 & 17.5032 \\
        FOMM~\cite{Siarohin2019nips} & 37.3648 & 0.7908 & 32.7782 & 3.3542 & 19.5126 \\
        \cmidrule(lr){1-6}
        \textbf{Ours} & \textbf{36.2377} & \textbf{0.8269} & \textbf{32.8591} & \textbf{3.4662} & \textbf{16.3049} \\ \bottomrule
        \end{tabular}
    }
  \end{minipage}
\end{table} 

\begin{table}[t]
  \begin{minipage}[t]{.48\linewidth}
    \centering
    \caption{Comparisons on {\mixamodata} in terms of MSE with the best results (lowest values) in bold.}
    \label{tab:mse-results-mixamo}
    \resizebox{\textwidth}{!}
    {
        \begin{tabular}{l|rrrr}
        \toprule
        Method & \textit{Andromeda} & \textit{Liam} & \textit{Remy} & \textit{Stefani} \\ \midrule
        NN & 24.9047 & 28.4846 & 27.1801 & 24.8609  \\
        \dif{CycleGAN~\cite{Zhu2017}} & 27.5520 & 29.9436 & 28.2237 & 22.7682  \\
        \dif{Pix2Pix~\cite{Isola2017}} & 23.9370 & 23.5610 & 24.0687 & 21.5841  \\
        \dif{EDN~\cite{Chan2019}} & 24.3244 & 23.1203 & 23.9229 & 35.0930  \\
        \dif{LWG~\cite{Liu2019}} & 24.2905 & 22.8587 & 22.9707 & 22.0910 \\
        \dif{MKN~\cite{Siarohin2019cvpr}} & 30.5934 & 39.3444 & 29.4817 & 24.1297 \\
        \dif{FOMM~\cite{Siarohin2019nips}} & 27.7809 & 29.0469 & 27.4474 & 25.3934 \\
        \cmidrule(lr){1-5}
        \textbf{Ours} & \textbf{23.8539} & \textbf{21.7328} & \textbf{22.0763} & \textbf{21.2587} \\ \bottomrule
        \end{tabular}
    }
    \end{minipage}
    \begin{minipage}[t]{.01\linewidth}
        \hfill
    \end{minipage}
    \hfill
    \begin{minipage}[t]{.48\linewidth}
    \centering
    \caption{Comparisons on {\mixamodata} in terms of SSIM with best results (highest values) in bold.}
    \label{tab:ssim-results-mixamo}
    \resizebox{0.95\textwidth}{!}
    {
        \begin{tabular}{l|rrrr}
        \toprule
        Method & \textit{Andromeda} & \textit{Liam} & \textit{Remy} & \textit{Stefani} \\ \midrule
        NN & 0.7357 & 0.7265 & 0.7487 & 0.7411 \\
        \dif{CycleGAN~\cite{Zhu2017}} & 0.7205 & 0.7154 & 0.7377 & 0.7753  \\
        \dif{Pix2Pix~\cite{Isola2017}} & 0.7784 & 0.7955 & 0.7932 & 0.8069  \\
        \dif{EDN~\cite{Chan2019}} & \textbf{0.7817} & 0.7931 & 0.7926 & 0.8058  \\
        \dif{LWG~\cite{Liu2019}} & 0.7613 & 0.7912 & 0.7887 & 0.7858  \\
        \dif{MKN~\cite{Siarohin2019cvpr}} & 0.7076 & 0.6874 & 0.7531 & 0.7753  \\
        \dif{FOMM~\cite{Siarohin2019nips}} & 0.7165 & 0.7404 & 0.7538 & 0.74642 \\
        \cmidrule(lr){1-5}
        \textbf{Ours} & 0.7726 & \textbf{0.8040} & \textbf{0.8057} & \textbf{0.8071}  \\ \bottomrule
        \end{tabular}
    }
    \end{minipage}
\end{table}

\subsubsection{Results on {\mixamodata}}
% Tables~\ref{tab:mse-results-mixamo},~\ref{tab:ssim-results-mixamo}~and~\ref{tab:psnr-results-mixamo} respectively show the quantitative results of pose transfer in terms of MSE, SSIM and PSNR on the {\mixamodata} dataset.
% Summarizing the aboving tables, Table~\ref{tab:ave-results-mixamo} shows the average quantitative results of pose transfer in terms of three metrics on {\mixamodata}.
\dif{
Tables~\ref{tab:mse-results-mixamo},~\ref{tab:ssim-results-mixamo},~\ref{tab:psnr-results-mixamo},~\ref{tab:IS-results-mixamo} and~\ref{tab:FID-results-mixamo} respectively show the quantitative results of pose transfer in terms of MSE, SSIM, PSNR, IS, and FID on the {\mixamodata} dataset.
Table~\ref{tab:ave-results-mixamo} provides comparisons on {\mixamodata} in terms of average results of the above five metrics over four characters.
}
We obtained the empirical results clearly demonstrated the effectiveness of {\ourmethod} on animation images:

\begin{itemize}
	\item \dif{
         As shown in Tables~\ref{tab:mse-results-mixamo},~\ref{tab:ssim-results-mixamo}, ~\ref{tab:psnr-results-mixamo},~\ref{tab:IS-results-mixamo} and~\ref{tab:FID-results-mixamo}, our {\ourmethod} outperformed all competing baselines regarding all five metrics on average again like the results on {\edndata}
        }
 %        As shown in Tables~\ref{tab:mse-results-mixamo},~\ref{tab:ssim-results-mixamo}~and~\ref{tab:psnr-results-mixamo}, our {\ourmethod} outperformed all competing baselines
	% regarding all three metrics on average again like the results on {\edndata}.	
	Notably, in terms of MSE, {\ourmethod} outperformed the second-best LWG by 0.8223 in Table~\ref{tab:mse-results-mixamo}.
	To be more concrete, the lowest MSEs of our {\ourmethod} indicated {\ourmethod} provided the most accurate motion transfer images with the shortest $L_2$ distance between the ground-truth images.
	For instance, our method provides 22.0763 of MSE on \textit{Remy} comparing to the 27.1801, 28.2237, 24.0687, and 23.9229 from NN, CycleGAN, Pix2Pix and EDN in the Table~\ref{tab:mse-results-mixamo}.
    \dif{Similar results of IS and FID can also be observed in Tables~\ref{tab:IS-results-mixamo} and~\ref{tab:FID-results-mixamo}.
    }
    Furthermore, together with the highest scores of PSNR and SSIM, our {\ourmethod} could generate synthesized images with the most accurate motion transfer and the best image quality simultaneously.
%     We can also make similar observations from \Cref{PSNR-results,SSIM-results}, where our method achieves the highest PSNR and SSIM scores on most characters.
% 	This observation shows that the synthesized images of our method have the best quality according to multiple metrics and demonstrates the effectiveness of our method.
        
	\item Secondly, NN provided a relative good results on \textit{Andromeda} but a higher MSE on \textit{Liam} and \textit{Remy} in the Table~\ref{tab:mse-results-mixamo},
    since its performance depends on the training dataset in the extreme, and it can't provide stable synthesized results.
	Moreover, in Tables~\ref{tab:mse-results-mixamo} and ~\ref{tab:ssim-results-mixamo}, CycleGAN also can not perform well, with the results are 27.5520, 29.9436, and 28.2237 MSE, and 0.7205, 0.7154 and 0.7377 for SSIM scores on \textit{Andromeda}, \textit{Liam}, and \textit{Remy} respectively.
	These results once again validated that it is difficult to learn a mapping from the skeleton images to human images directly using unpaired data.
    % Secondly, we notice CycleGAN has the worst SSIM results in the \Cref{tab:ssim-results-mixamo}.
    % We argue that CycleGAN is better at transferring the color or style for images from two domains rather than change the geometry of the images, such as recover the human appearance from the human skeleton, because CycleGAN aims to learn a paired mapping from the skeleton images to human images directly.
    % This property of CycleGAN is also supported by its inferior PSNR scores comparing with other methods.
    % For example, we can see that the SSIM score of CycleGAN on Andromeda is 0.7205 comparing with 0.7357, 0.7784, 0.7817 and 0.7726 from NN, Pix2Pix, EDN and our method.
    
    \item \dif{
    Thirdly, from Tables~\ref{tab:mse-results-mixamo},~\ref{tab:ssim-results-mixamo},~\ref{tab:psnr-results-mixamo},~\ref{tab:IS-results-mixamo}, and~\ref{tab:FID-results-mixamo}, Pix2Pix, EDN, and LWG delivered such superior performance comparing with NN and CycleGAN.
    }
    % Thirdly, from the Tables~\ref{tab:mse-results-mixamo},~\ref{tab:ssim-results-mixamo},~\ref{tab:psnr-results-mixamo}, Pix2Pix, EDN and LWG delivered such superior performance comparing with NN and CycleGAN.
    because they aim to learn a paired mapping from the skeleton images to human images directly.
    Besides that, the skeleton images as their input have accurate motion information without any noise and make \dif{the} learning process easier.
    In contrast, even without converting source person images to skeleton images, our method {\ourmethod} fills the gap between \dif{the} original source person images and the skeleton images to some extent by introducing the keypoint amplifier, two consistency losses, and a support dataset for training.
    Without any pre-process steps, and thus the source person images contain much noise and redundant information, 
    % our method can still achieve more improvements over Pix2Pix and EDN according to all 3 metrics.
    \dif{
    our method can still achieve more improvements over Pix2Pix and EDN according to all five metrics.
    }
    % our method can achieve tremendous improvements on Pix2Pix and EDN since the input of our method are the original source person images without any pre-process steps and thus contains much noise and redundant information.
    
    % \item  Pix2Pix, EDN and LWG delivered superior performance comparing with NN and CycleGAN. 
    % Good results are also provided by Pix2Pix and EDN of 4 characters from both \Cref{MSE-results} and \Cref{LPIPS-results}.
    
	\item Compared to the results on {\edndata}, MKN and FOMM dropped their performance when they transferred the pose between different people on {\mixamodata}.
	It is difficult for them to extract keypoints features without a pre-trained pose estimator when the source person was not in the training dataset.
	For example, in Table~\ref{tab:ssim-results-mixamo}, we can see that the SSIM score of MKN on Andromeda is 0.7076 \dif{compared} with 0.7357, 0.7784, 0.7817 and 0.7726 from NN, Pix2Pix, EDN, and our method.
\end{itemize}

\begin{table}[t]
  \begin{minipage}[t]{.48\linewidth}
    \centering
    \caption{Comparisons on {\mixamodata} in terms of PSNR with the best results (highest values) in bold.}
    \label{tab:psnr-results-mixamo}
    \resizebox{\textwidth}{!}
    {
        \begin{tabular}{l|rrrr}
        \toprule
        Method & \textit{Andromeda} & \textit{Liam} & \textit{Remy} & \textit{Stefani} \\ \midrule
        NN & \textbf{34.4420} & 34.0575 & 34.3865 & 34.3977  \\
        \dif{CycleGAN~\cite{Zhu2017}} & 33.7903 & 33.4230 & 33.6786 & 34.6130  \\
        \dif{Pix2Pix~\cite{Isola2017}} & 34.4049 & 34.5167 & 34.3875 & 34.8497  \\
        \dif{EDN~\cite{Chan2019}} & 34.3524 & 34.5882 & 34.4214 & 34.7247  \\
        \dif{LWG~\cite{Liu2019}} & 34.3426 & 34.6762 & 34.6183 & 34.7519  \\
        \dif{MKN~\cite{Siarohin2019cvpr}} & 33.6626 & 32.8841 & 33.7487 & 34.4779 \\
        \dif{FOMM~\cite{Siarohin2019nips}} & 33.7777 & 33.5912 & 33.8337 & 34.1919  \\
        \cmidrule(lr){1-5}
        \textbf{Ours} & 34.4336 & \textbf{34.8798} & \textbf{34.7823} & \textbf{34.9303} \\ \bottomrule
        \end{tabular}
    }
    \end{minipage}
    \begin{minipage}[t]{.01\linewidth}
        \hfill
    \end{minipage}
        \hfill
    \begin{minipage}[t]{.48\linewidth}
    \centering
    \caption{\dif{Comparisons on {\mixamodata} in terms of IS with the best results (highest values) in bold.}}
    \label{tab:IS-results-mixamo}
    \resizebox{0.94\textwidth}{!}
    {
        \begin{tabular}{l|rrrr}
        \toprule
        Method & \textit{Andromeda} & \textit{Liam} & \textit{Remy} & \textit{Stefani} \\ \midrule
        NN & 3.3671 & 3.4907 & 3.3170 & 3.3815 \\
        CycleGAN~\cite{Zhu2017} & 2.9237 & 2.9105 & 3.0148 & 3.0681 \\
        Pix2Pix~\cite{Isola2017} & 3.3892 & 3.5026 & 3.3356 & 3.5073  \\
        EDN~\cite{Chan2019} & \textbf{3.4308} & 3.5418 & 3.4509 & 3.5624 \\
        LWG~\cite{Liu2019} & 3.3704 & 3.4052 & 3.3177 & 3.4339 \\
        MKN~\cite{Siarohin2019cvpr} & 3.3856 & 3.5697 & 3.4212 & 3.5105  \\
        FOMM~\cite{Siarohin2019nips} & 3.3921 & 3.4893 & 3.4082 & 3.4721  \\
        \cmidrule(lr){1-5}
        \textbf{Ours} & 3.4285 & \textbf{3.6297} & \textbf{3.5618} & \textbf{3.6075} \\ \bottomrule
        \end{tabular}
    }
  \end{minipage}
\end{table} 

\begin{table}[t]
  \begin{minipage}[t]{.48\linewidth}
    \centering
    \caption{\dif{Comparisons on {\mixamodata} in terms of FID with the best results (lowest values) in bold.}}
    \label{tab:FID-results-mixamo}
    \resizebox{0.92\textwidth}{!}
    {
        \begin{tabular}{l|rrrr}
        \toprule
        Method & \textit{Andromeda} & \textit{Liam} & \textit{Remy} & \textit{Stefani} \\ \midrule
        NN & 21.9411 & 23.4728 & 22.5086 & 24.7382  \\
        CycleGAN~\cite{Zhu2017} & 23.1479 & 25.1871 & 21.0418 & 21.2344 \\
        Pix2Pix~\cite{Isola2017} & 14.5051 & 12.7375 & 12.8751 & 11.1583  \\
        EDN~\cite{Chan2019} & 14.6281 & 11.1481 & 11.6892 & 11.2089 \\
        LWG~\cite{Liu2019} & 16.7303 & 11.4930 & 13.5728 & 14.3207 \\
        MKN~\cite{Siarohin2019cvpr} & 21.4839 & 21.4015 & 17.3208 & 16.7219  \\
        FOMM~\cite{Siarohin2019nips} & 19.6782 & 19.3755 & 21.2926 & 20.0382  \\
        \cmidrule(lr){1-5}
        \textbf{Ours} & \textbf{14.3172} & \textbf{10.1062} & \textbf{11.2756} & \textbf{10.7603} \\ \bottomrule
        \end{tabular}
    }
    \end{minipage}
    \begin{minipage}[t]{.01\linewidth}
        \hfill
    \end{minipage}
        \hfill
    \begin{minipage}[t]{.48\linewidth}
    \centering
    \caption{\dif{Comparisons on {\mixamodata} in terms of the average results of the 5 metrics over 4 characters.}}
    \label{tab:ave-results-mixamo}
    \resizebox{\textwidth}{!}
    {
        \begin{tabular}{l|rrrrr}
        \toprule
        Method & MSE($\downarrow$) & SSIM($\uparrow$)  & PSNR($\uparrow$) & IS($\uparrow$) & FID($\downarrow$) \\ \midrule
        NN & 26.3576 & 0.7380 & 34.3209 & 3.3891 & 23.1652 \\
        CycleGAN~\cite{Zhu2017} & 27.1219 & 0.7372 & 33.8762 & 2.9793 & 22.6528 \\
        Pix2Pix~\cite{Isola2017} & 23.2877 & 0.7935 & 34.5397 & 3.4337 & 12.8190 \\
        EDN~\cite{Chan2019} & 26.6151 & 0.7933 & 34.5217 & 3.4965 & 12.1686 \\
        LWG~\cite{Liu2019} & 23.0527 & 0.7817 & 34.5972 & 3.3818 & 14.0292 \\
        MKN~\cite{Siarohin2019cvpr} & 30.8873 & 0.7308 & 33.6933 & 3.4718 & 19.2320 \\
        FOMM~\cite{Siarohin2019nips} & 28.6544 & 0.7392 & 33.8486 & 3.4404 & 20.0961 \\
        \cmidrule(lr){1-6}
        \textbf{Ours} & \textbf{22.2304} & \textbf{0.7973} & \textbf{34.7565} & \textbf{3.5569} & \textbf{11.6148} \\ \bottomrule
        \end{tabular}
    }
  \end{minipage}
\end{table}

% \begin{table}[t]
% \centering
% \resizebox{0.7\columnwidth}{!}
% {\begin{tabular}{l|rrrr|r}
% \toprule
% Method & \textit{Andromeda} & \textit{Liam} & \textit{Remy} & \textit{Stefani} & Average \\ \midrule
% NN & \textbf{34.4420} & 34.0575 & 34.3865 & 34.3977 & 34.3209 \\
% CycleGAN~\cite{Zhu2017} & 33.7903 & 33.4230 & 33.6786 & 34.6130 & 33.8762 \\
% Pix2Pix~\cite{Isola2017} & 34.4049 & 34.5167 & 34.3875 & 34.8497 & 34.5397 \\
% EDN~\cite{Chan2019} & 34.3524 & 34.5882 & 34.4214 & 34.7247 & 34.5217 \\
% LWG~\cite{Liu2019} & 34.3426 & 34.6762 & 34.6183 & 34.7519 & 34.5972 \\
% MKN~\cite{Siarohin2019cvpr} & 33.6626 & 32.8841 & 33.7487 & 34.4779 & 33.6933 \\
% FOMM~\cite{Siarohin2019nips} & 33.7777 & 33.5912 & 33.8337 & 34.1919 & 33.8486 \\
% \cmidrule(lr){1-6}
% \textbf{Ours} & 34.4336 & \textbf{34.8798} & \textbf{34.7823} & \textbf{34.9303} & \textbf{34.7565} \\ \bottomrule
% \end{tabular}}
% \caption{Comparisons on {\mixamodata} in terms of PSNR with the best results (highest values) in bold.}
% \label{tab:psnr-results-mixamo}
% \end{table}

\subsection{Ablation Study}
\label{sec:abl_study}
\label{sec:ablation}

To better understand the merits of designs of {\ourmethod}, we conducted detailed ablation studies on \textit{Subject1} from {\edndata} and \textit{Liam} from {\mixamodata}.
The evaluation results are shown in Tables~\ref{tab:ab_study-edn}~and~\ref{tab:ab_study-mixamo}.

\subsubsection{Keypoint Amplifier}
In Tables~\ref{tab:ab_study-edn}~and~\ref{tab:ab_study-mixamo}, the baseline (the first row) is the results of our model without the keypoint amplifier, the consistency losses and the support dataset.
This baseline provided the worst results of the three metrics, e.g., the MSE is 26.4595 in Table~\ref{tab:ab_study-mixamo}.
The results in the second row versus those in the first row showed that the keypoint amplifier strengthened the performance of pose transfer, such as promoting the SSIM of the baseline from 0.6595 to 0.6676 in Table~\ref{tab:ab_study-edn}.
Even with the consistency losses and support set (rows 6 and 7), it continuously
reduced the interference of the noise on the keypoint heatmaps.
These results validated that the keypoint amplifier can filter out the real keypoint locations and reduce the interference of the noise on the keypoint heatmaps.

% At first, The baseline (the first row) is the results of our model without the keypoint amplifier,
% the consistency losses and the support dataset.
% It provides the worst results of the 4 metrics,
% e.g, the MSE is 26.4595.
% For the second row,
% we introduce a keypoint amplifier which makes a tremendous improvement on all the 4 metrics such as promoting the SSIM of the baseline from 0.7605 to 0.7726.
% We believe that the keypoint amplifier would filter out the real keypoint locations and reduce the interference of the noise on the keypoint heatmaps.
% \paragraph{Consistency losses}

\begin{table}[t]
    \begin{minipage}[t]{.48\linewidth}
    \centering
    \caption{Ablation studies on \textit{Subject1} from {\edndata}. \textit{KA} denotes the {\ka}.}
    \label{tab:ab_study-edn}
    \resizebox{\textwidth}{!}
    {
        \begin{tabular}{c|cccc|ccc}
        \toprule
        &\textit{KA} & $\mL_\rsc$ & $\mL_\rmc$ & $\mL_\rsup$ & MSE($\downarrow$) & PSNR($\uparrow$)  & SSIM($\uparrow$) \\ \midrule
        1&  &   &   &   & 52.5749 & 30.9361 & 0.6595 \\
        2&\ding{51} &   &   &   & 50.7810 & 31.0860 & 0.6676 \\
        3&\ding{51} & \ding{51} &   &   & 49.3831 & 31.2043 & 0.6742 \\
        4&\ding{51} &   & \ding{51} &   & 49.8352 & 31.1662 & 0.6710 \\
        5&\ding{51} & \ding{51} & \ding{51} &   & 48.6431 & 31.2714 & 0.6796 \\
        6& & \ding{51} & \ding{51} & \ding{51} & 46.3699 & 31.4801 & 0.6937 \\
        \cmidrule(lr){1-8}
        7&\ding{51} & \ding{51} & \ding{51} & \ding{51} & \textbf{45.2043} & \textbf{31.5978} & \textbf{0.7083} \\ \bottomrule
        \end{tabular}
    }
    \end{minipage}
    \begin{minipage}[t]{.01\linewidth}
        \hfill
    \end{minipage}
    \hfill
    \begin{minipage}[t]{.48\linewidth}
    \centering
    \caption{Ablation studies on \textit{Liam} from {\mixamodata}. \textit{KA} denotes the {\ka}.}
    \label{tab:ab_study-mixamo}
    \resizebox{\textwidth}{!}
    {
        \begin{tabular}{c|cccc|ccc}
        \toprule
        &\textit{KA} & $\mL_\rsc$ & $\mL_\rmc$ & $\mL_\rsup$ & MSE($\downarrow$) & PSNR($\uparrow$)  & SSIM($\uparrow$) \\ \midrule
        1&  &   &   &   & 26.4595 & 33.9856 & 0.7605 \\
        2&\ding{51} &   &   &   & 24.7312 & 34.2838 & 0.7726 \\
        3&\ding{51} & \ding{51} &   &   & 23.7580 & 34.4666 & 0.7830 \\
        4&\ding{51} &   & \ding{51} &   & 22.6160 & 34.6913 & 0.8011 \\
        5&\ding{51} & \ding{51} & \ding{51} &   & 21.9509 & 34.8292 & 0.8032 \\
        6&  & \ding{51} & \ding{51} & \ding{51}  & 22.0240 & 34.8047 & 0.8031 \\
        \cmidrule(lr){1-8}
        7&\ding{51} & \ding{51} & \ding{51} & \ding{51} & \textbf{21.7328} & \textbf{34.8798} & \textbf{0.8040} \\ \bottomrule
        \end{tabular}
    }
    \end{minipage}
\end{table}

\subsubsection{Consistency Losses}
We explored the effects of consistency losses $\mL_\rsc$ and $\mL_\rmc$ on the task of pose transfer.
Considering the different results between the second row and the third row, the static feature consistency loss $\mL_\rsc$ boosted the pose transfer task and verified its validity. 
Moreover, the performance variations between the second row and the fourth row clearly evidenced the advantages of pose feature consistency loss $\mL_\rmc$.
Together with these two feature consistency losses, the model achieved better results with MSE of 21.9509, PSNR of 34.8292, and SSIM of 0.8032, as shown in the fifth row of Table~\ref{tab:ab_study-mixamo}
The comparison among these cases showed that each consistency loss, whether the static one or the motion one, can enforce the consistency between the real image and synthesized images,
thus \dif{improving} the quality of the synthesized images.

It is \dif{noted} that the static feature consistency was more useful than the motion one on {\edndata}, while we observed the opposite performance on {\mixamodata}. 
For example, in the third and fourth rows of Table~\ref{tab:ab_study-mixamo}, the static feature consistency loss achieved 0.0104 improvements for SSIM,
while the motion feature consistency loss provided \dif{a} 0.0285 improvement.
We considered that since the backgrounds of images in {\mixamodata} were simply white (i.e., easy to learn), the static feature consistency could only boost the performance for personal appearance on {\mixamodata}.
On the contrary, the backgrounds were more complex in {\edndata}. Thus the static feature consistency was more effective on {\edndata} than on {\mixamodata}.

% \todo{WK: it seems that motion is more useful than static on \mixamodata, and the opposite on \edndata. Could you validate and discuss this?}

% Together with these two feature consistency losses, the model achieved better results with MSE of 21.9509, PSNR of 34.8292 and SSIM of 0.8032, as shown in the fifth row of Table~\ref{tab:ab_study-mixamo}
% Second, we introduce the static feature and motion feature consistency losses for the third and fourth rows respectively.
% We note that the improvements from the motion feature consistency loss is more than the improvements from the static feature consistency loss.
% For example, the static feature consistency loss achieves 0.0104 improvement for SSIM,
% while the motion feature consistency loss provides 0.0285 decrease.
% When these two consistency losses (fifth row) are both added,
% the results for all the metrics are better than the cases with only one consistency loss, like reducing the MSE from 23.7580 and 22.6160 to 21.9509.
% The comparison among these cases showed that each consistency loss whether the static one or the motion one can enforce the consistency between the real image and synthesized images,
% thus improve the quality of the synthesized images.

\subsubsection{Support Set with Augmented Consistency Loss}
% Finally, 
We added the support set and utilized the augmented consistency loss $\mL_\rsup$ during the training. 
The comparisons among the final row and other rows notably indicated that the support set could improve the generalization ability and the robustness of our model, especially when the poses of the source person are closer to the poses in the support set.
The support set provided more negative examples besides the original training set and could help the discriminator to form an accurate boundary, which could further strengthen the performance of the generator.
We also noticed that the support set was more useful on {\edndata}, because {\edndata} was sampled from the video clips of subjects, and it did not contain much motion variance compared with {\mixamodata}. 
Moreover, the limb ratios and the distances between the person and camera in \dif{the} support set were also distinct from {\edndata}, which provided more valuable negative samples for {\edndata}.

\dif{
\subsubsection{Different Pose Estimator Backbone}

To further verify the contribution of different pre-trained pose estimator backbones,
we also conducted extensive experiments on \textit{Subject1} from {\edndata} and \textit{Liam} from {\mixamodata} with ResNet architectures including ResNet-18, ResNet-50, ResNet-101, and ResNet-152.
For ResNet-18, we did not find any pre-trained model online, and thus we re-implemented~\cite{Cao2017}  by replacing the VGG-19 with the ResNet-18 following the same training process as~\cite{Cao2017}.
For ResNet-50, ResNet-101, and ResNet-152, we directly use the pre-trained models from~\cite{Xiao2018simple}.
The results are shown in Tables~\ref{tab:ab_study-backbone-edn}~and~\ref{tab:ab_study-backbone-mixamo}.
We can observe that when we use a larger pose estimator backbone, we usually get better performance according to all five metrics.
For instance, on \textit{Subject1} from {\edndata}, the ResNet-152 achieved the best SSIM result of 0.7261 compared to other backbones.
It is because a better pre-trained pose estimator backbone can provide more representative pose information and thus generate high-fidelity images.
It suggests that our method still has the potential for improvement by combining more advanced pose estimation technologies.
In this work, we mainly focus on how to add consistency in the feature space instead of directly using a better pose estimator backbone.
}

\subsection{Qualitative Results}
\label{sec:qualitative}
\begin{table}[t]
  \begin{minipage}[t]{.48\linewidth}
    \centering
    \caption{\dif{Ablation studies of different pose estimator backbones on \textit{Subject1} from {\edndata}.}}
    \label{tab:ab_study-backbone-edn}
     \resizebox{\textwidth}{!}
    {
        \begin{tabular}{c|ccccc}
        \toprule
        Network & MSE($\downarrow$) & PSNR($\uparrow$)  & SSIM($\uparrow$) & IS($\uparrow$) & FID($\downarrow$) \\ \midrule
        ResNet-18 & 49.3291 & 30.5388 & 0.6802 & 3.2811 & 20.7048 \\
        ResNet-50~\cite{Xiao2018simple} & 44.3875 & 31.7714 & 0.7039 & 3.3255 & 18.1374 \\
        ResNet-101~\cite{Xiao2018simple} & 42.8027 & 31.9208 & 0.7255 & 3.4039 & 16.8275 \\
        ResNet-152~\cite{Xiao2018simple} & \textbf{41.5562} & \textbf{32.3248} & \textbf{0.7261} & \textbf{3.4415} & \textbf{16.4553} \\
        \cmidrule(lr){1-6}
        VGG-19~\cite{Cao2017} & 45.2043 & 31.5978 & 0.7083 & 3.3502 & 18.3029  \\ \bottomrule
        \end{tabular}
    }
    \end{minipage}
    \begin{minipage}[t]{.01\linewidth}
        \hfill
    \end{minipage}
    \hfill
    \begin{minipage}[t]{.48\linewidth}
    \centering
    \caption{\dif{Ablation studies of different pose estimator backbones on \textit{Liam} from {\mixamodata}.}}
    \label{tab:ab_study-backbone-mixamo}
     \resizebox{\textwidth}{!}
    {
        \begin{tabular}{c|ccccc}
        \toprule
        Network & MSE($\downarrow$) & PSNR($\uparrow$)  & SSIM($\uparrow$) & IS($\uparrow$) & FID($\downarrow$) \\ \midrule
        ResNet-18 & 23.3403 & 34.3123 & 0.8057 & 3.5564 & 10.2478 \\
        ResNet-50~\cite{Xiao2018simple} & 21.3634 & 35.2125 & 0.8073 & 3.6187 & 9.9064 \\
        ResNet-101~\cite{Xiao2018simple} & 20.8277 & 35.8716 & \textbf{0.8342} & 3.6571 & 9.5731 \\
        ResNet-152~\cite{Xiao2018simple} & \textbf{20.5169} & \textbf{35.8920} & 0.8188 & \textbf{3.6933} & \textbf{9.4565}  \\
        \cmidrule(lr){1-6}
        VGG-19~\cite{Cao2017} & 21.7328 & 34.8798 & 0.8040 & 3.6297 & 10.1062 \\ \bottomrule
        \end{tabular}
    }
    \end{minipage}
\end{table} 

\begin{figure*}[t]
    \centering
    \includegraphics[width=0.98\linewidth]{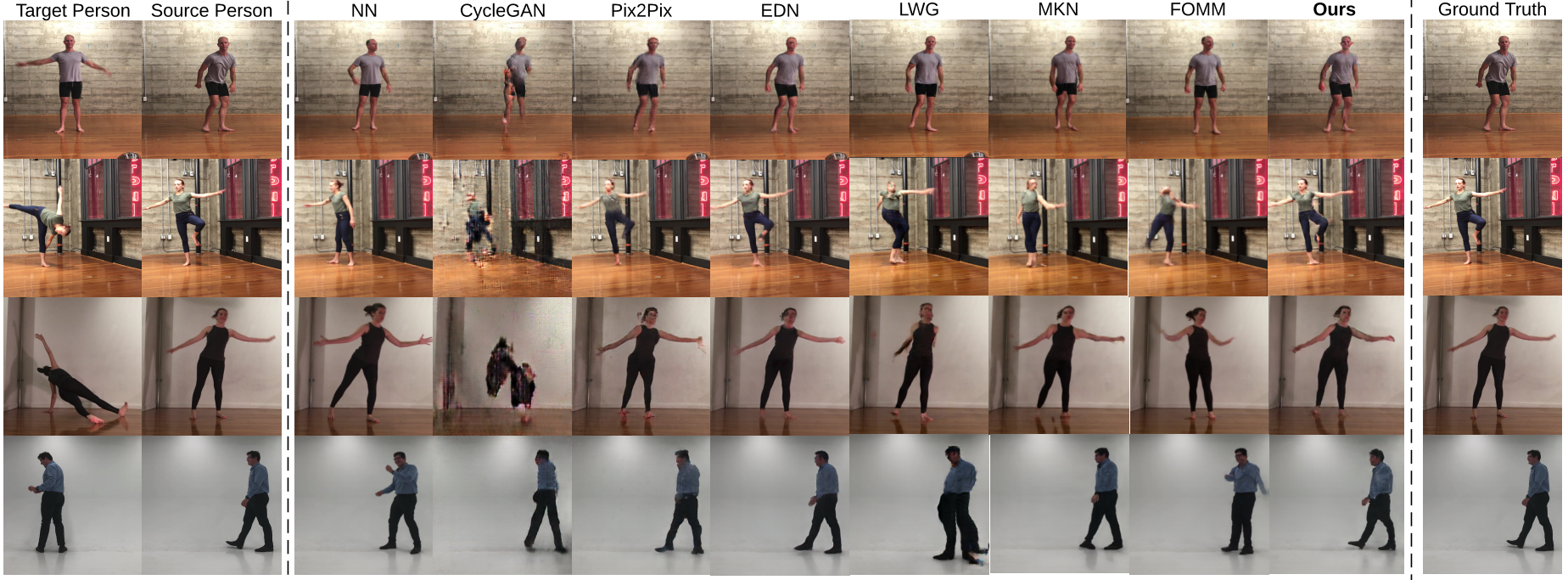}
    \caption{Visualizations of pose transfer with different characters and various poses on \edndata. The columns from left to right show target persons, source persons, results of different methods, and ground truth, respectively.}
    \label{Fig:MT-main-results-edn}
\end{figure*}

\begin{figure*}[t]
    \centering
    \includegraphics[width=0.98\linewidth]{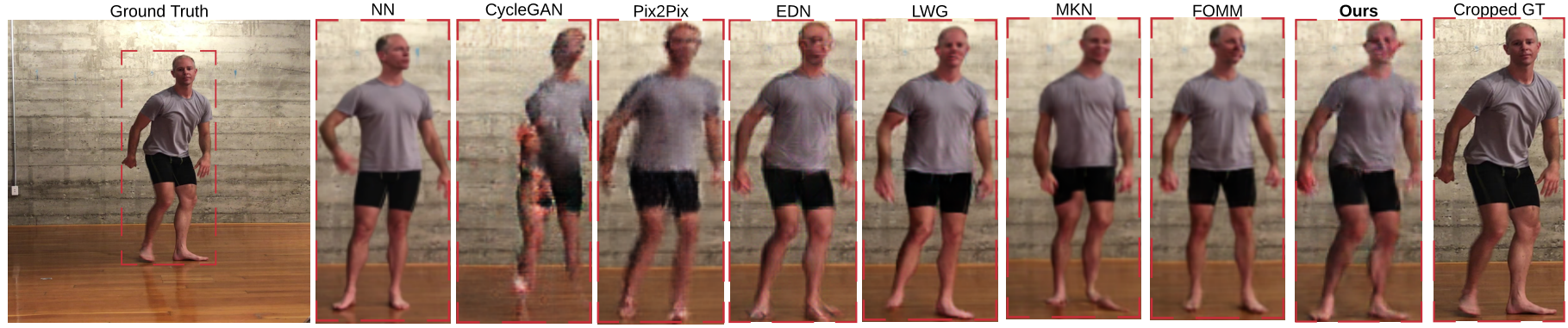}
    \caption{Visualizations of \textit{Subject1} from {\edndata} with more details. The image in the first left column is the source image and ground truth. We did not show the target image (i.e, the same person with different pose) due to space limitation and cropped the results to the areas surrounded by the red dash lines.}
    \label{Fig:MT-sub1-crop-results-edn}
%\vspace{-0.1cm}
\end{figure*}

\begin{figure*}[t]
    \centering
    \includegraphics[width=0.98\linewidth]{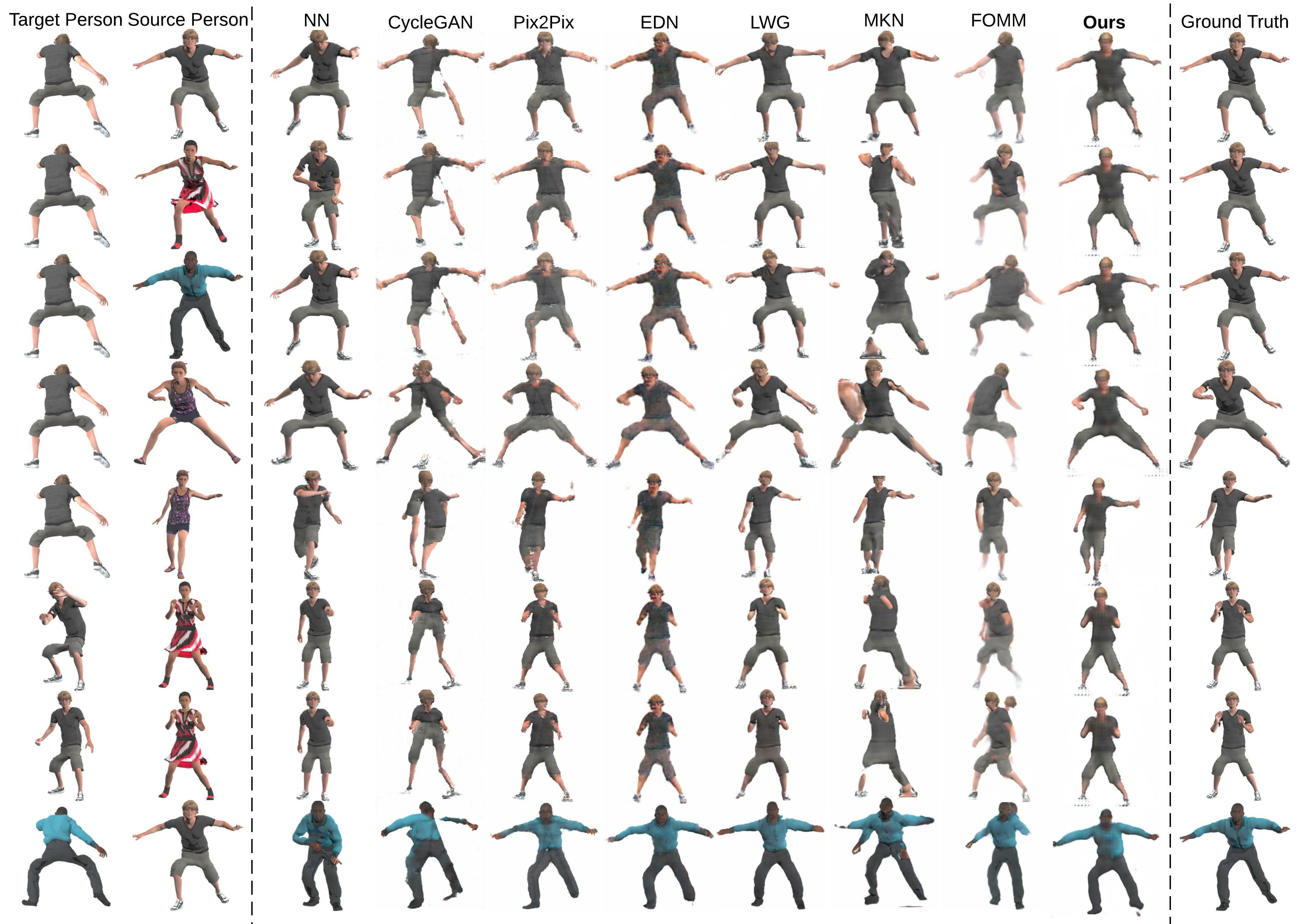}
    \caption{Visualizations of motion transfer with different characters and various motions on \mixamodata. The columns from left to right show target persons, source persons, results of different methods, and ground truth, respectively.}
    \label{Fig:MT-main-results-mixamo}
\end{figure*}

% \begin{figure*}[t]
%     \centering
%     \includegraphics[width=0.98\linewidth]{Figure/EDN-sub1-MT-abl-results.jpeg}
%     \caption{Visualizations of \textit{Subject1} from {\edndata}. The columns from left to right show target persons, source persons, results of 6 different settings in ablation studies, and ground truth, respectively.
%     The indices of each column are the same as in the Table~\ref{tab:ab_study-mixamo}.}
%     \label{Fig:MT-sub1-abl-results-end}
% \end{figure*}

\begin{figure*}[t]
    \centering
    \includegraphics[width=0.98\linewidth]{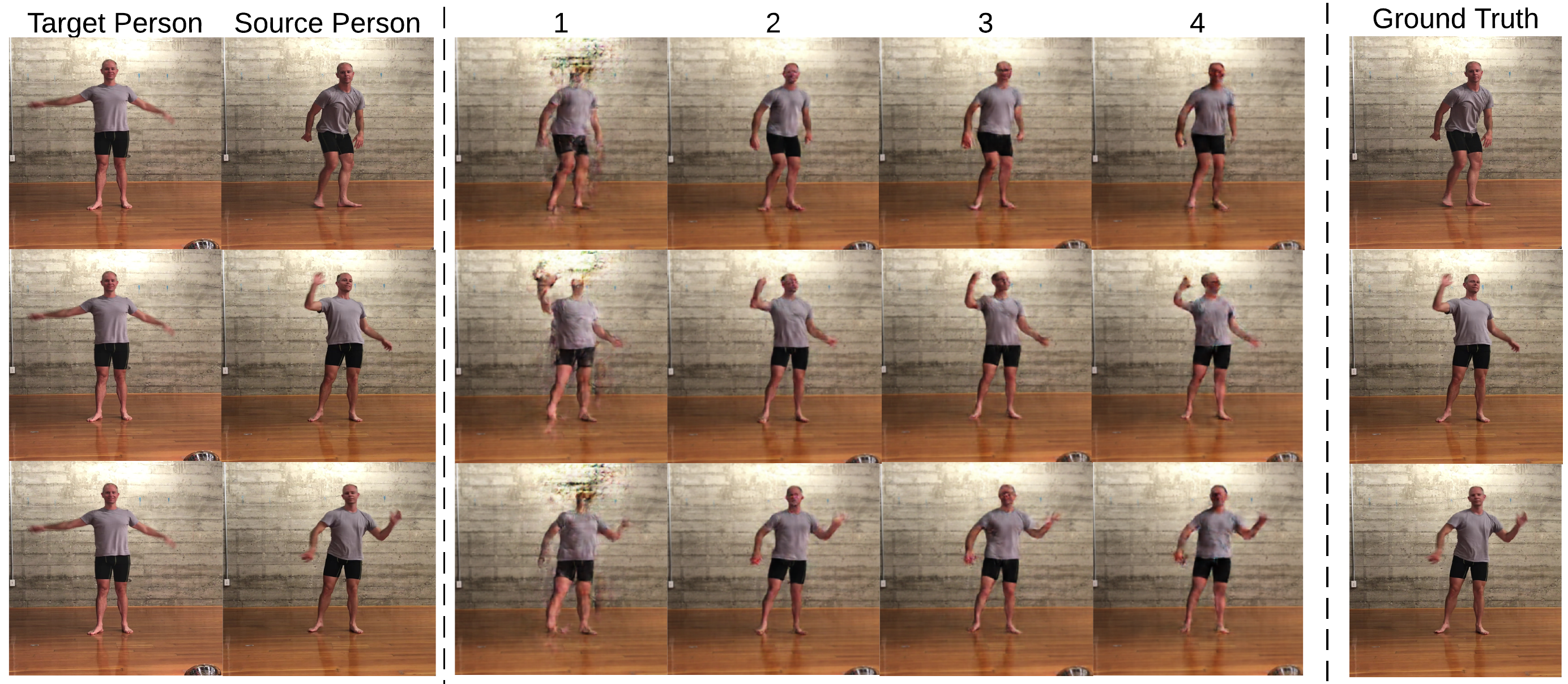}
    
    \includegraphics[width=0.98\linewidth]{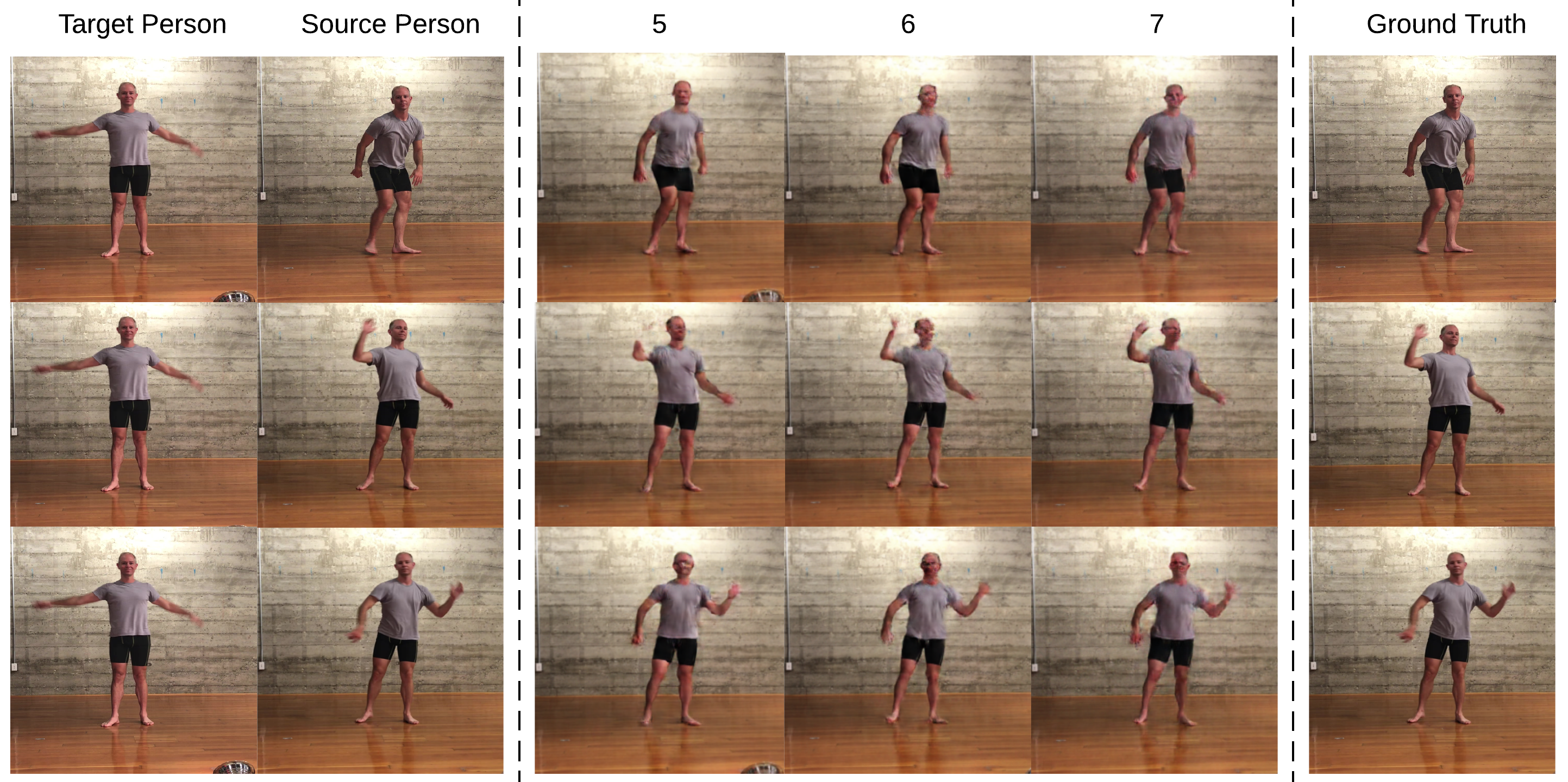}
    \caption{\dif{Visualizations of \textit{Subject1} from {\edndata}. The columns from left to right show target persons, source persons, results of 6 different settings in ablation studies, and ground truth, respectively.
    The indices of each column are the same as in the Table~\ref{tab:ab_study-mixamo}}.}
    \label{Fig:MT-sub1-abl-results-end}
\end{figure*}

We further highlighted the superiority of our proposed approach by showing and contrasting the visualizations of synthesized results of all models on {\edndata} and {\mixamodata}, respectively.
In addition, we presented the visualizations of the ablation comparisons for \textit{Subject1} on {\edndata} to indicate the effectiveness of components of {\ourmethod}.

\subsubsection{Visual comparisons on {\edndata}}
We also provided visualization results on {\edndata} in Figure~\ref{Fig:MT-main-results-edn} and cropped the results in Figure~\ref{Fig:MT-sub1-crop-results-edn} with more details, and {\ourmethod} synthesized real person image results with better qualities, which clearly indicated the effectiveness of our method as well.
For instance:

\begin{itemize}
    \item As shown in the second row of Figure~\ref{Fig:MT-main-results-edn}, the difference between the source pose and the target pose is very large, and this posture involves changes in almost all parts of the human body.
    In that case, we can observe that {\ourmethod} can generate the correct human pose for the \textit{Subject2}, when other methods like LWG, MKN, and FOMM can only synthesize distorted poses.
    Another example is in Figure~\ref{Fig:MT-sub1-crop-results-edn}, we observed that the synthesized images of our method contained more details, including the clear face, wrinkles on clothes, etc.
    The generated pose transfer image of \textit{Subject1} from {\ourmethod} has a more clear face structure, and the face orientation of the characters is also consistent with the target image, while MKN and FOMM failed to synthesize correct face orientation and hand poses.
    
 	\item Sometimes NN could generate images that are closer to the ground truths, e.g., \dif{the} image of \textit{Subject3} at the third row of Figure~\ref{Fig:MT-main-results-edn}, since the number of images in the training set increases from 1488 in {\mixamodata} to 10000 in {\edndata}. But people could still easily determine that the poses of the generated images are different from the ground-truths.
 	
 	\item The results produced by Pix2Pix and EDN were more blurry, and there was an aliasing effect on the edges of the subject in the Pix2Pix result. 
 	In addition, EDN showed better image results than Pix2Pix did. In the third row of Figure~\ref{Fig:MT-main-results-edn}, we noticed that the image of \textit{Subject3}, synthesized by Pix2Pix, had much noise in the subject's hair. On the contrary, the result of EDN was smoother and more realistic, but they lacked many details compared with {\ourmethod}, especially the face part was still twisted.
\end{itemize}

% \begin{figure*}[t]
%     \centering
%     \includegraphics[width=0.82\linewidth]{Figure/EDN-sub1-MT-crop-results.png}
%     \caption{Visualizations of \textit{Subject1} from {\edndata} with more details. We cropped the results to the areas surrounded by the red dash lines.}
%     \label{Fig:MT-sub1-crop-results-edn}
% %\vspace{-0.1cm}
% \end{figure*}

\subsubsection{Visual comparisons on {\mixamodata}}
As depicted in Figure~\ref{Fig:MT-main-results-mixamo}, we employed multiple target persons and source persons with various poses that were different from those in the training set of the {\mixamodata}.
The first column is the target person image, and the second one is the source person image.
Note that only our method {\ourmethod} takes the target person and source person images at the same time.
CycleGAN, Pix2Pix, and EDN only take the corresponding skeleton image of the source person image as input and synthesize results.
The last column is real images of the target person \dif{making} the desired motion which are ground truth.
% Compared with the aforementioned baselines, {\ourmethod} offered better transfer results and alleviated their drawbacks on synthesizing body parts, clothes, faces, just name a few. 
We illustrate details as follows:

\begin{itemize}
    \item Firstly, compared to the aforementioned baselines, {\ourmethod} offered better transfer results and alleviated their drawbacks on synthesizing body parts, clothes, \dif{and} faces, just \dif{to} name a few.
    For instance, only {\ourmethod} generated the left hand of \textit{Liam} at the first and second rows, while there \dif{was} only the arm in \dif{the} results of other baselines.
    When the target person and source person are not the same person, e.g., in rows 2-8, {\ourmethod} can still disentangle the pose information of the source person from the static information successfully and then synthesize high-quality results. 
    In contrast, MKN and FOMM are unable to strip out the pose information fully, and thus the generated results are strange, e.g., the person in the result of FOMM is facing away in the fourth row.
    
    \item Meanwhile, though the Pix2Pix and EDN could synthesize the target persons with the desired poses, some body parts (e.g., arms, hands) generated by Pix2Pix were severely corrupted.
    For example, images produced by EDN had stale colors and lots of noisy pixels in terms of clothes and faces, and it failed to capture the details of shoes \dif{in} the 3rd row since the color of \dif{the} shoes is white and is easily confused with the background. 
 	
 	\item Moreover, even though the image results of NN are more clear and not blurry, the NN method always made incorrect pose transfer since it can solely synthesize poses \dif{that} existed in the training set.
 	Besides that, NN can not synthesize images with temporal coherence, while the input is usually a sequence of desirable motions rather than a single image.
 	
 	\item The CycleGAN could not provide similar poses compared with the ground truths.
 	We can observe that almost all the poses were distorted, and the qualities of \dif{the} corresponding images were constantly degraded.	
\end{itemize}

\subsubsection{Visual ablation studies on \textit{Subject1}}
We presented the visualization of the ablation study results on \textit{Subject1} in Figure~\ref{Fig:MT-sub1-abl-results-end}. \dif{We} observed that each component has different degrees of improvement to the results, including clear head, hand gestures, etc. Specifically:

\begin{itemize}
    \item The results produced by setting 1, which does not have the Keypoint Amplifier, have \dif{more} noise than other settings, especially near the head area.
    Compared to the baseline setting 1, setting 2 with Keypoint Amplifier significantly improved the generated image qualities.
    For instance, the results of setting 1 in the first and second rows failed to synthesize complete head and hands, while the corresponding results of setting 2 fixed these errors.
    \item As shown in the column of setting 7, {\ourmethod} with full components synthesized images with more accurate details and better qualities. 
    In the second row of the results produced by setting 5, we can see the generated person lacked his arm, while the results of setting 7 with the augmented consistency loss from \dif{the} support set can complete the arm part. 
\end{itemize}

\section{Conclusion}
\label{sec:conclusion}
In this paper, we proposed {\ourmethod}, a novel network with disentangled feature consistencies for human pose transfer.
We introduce two disentangled feature consistency losses to enforce the pose and static information to be consistent between the synthesized and real images.
Besides, we leverage the keypoint amplifier to denoise the keypoint heatmaps and make it easier to extract the pose features.
Moreover, we show that the support set {\supdata} containing different subjects with unseen poses can boost the pose transfer performance and enhance the robustness of the model.
To enable the accurate evaluation of pose transfer between different people, we collect an animation character dataset {\mixamodata}.
Results on both animation and real image datasets, {\mixamodata} and {\edndata}, consistently demonstrated the effectiveness of the proposed model.

\bibliographystyle{ACM-Reference-Format}
\bibliography{reference}

\end{document}